\documentclass[journal]{IEEEtran}
\usepackage{multirow}
\usepackage{booktabs}
\usepackage{makecell}
\usepackage{cite}
\usepackage{graphicx}
\usepackage{amsmath}
\usepackage{url}
\usepackage{mathrsfs}
\usepackage{bm}
\usepackage{caption}
\usepackage[ruled,vlined]{algorithm2e}
\usepackage[colorlinks,linkcolor=blue]{hyperref}
\usepackage{color}
\usepackage{amssymb}
\usepackage{subfigure}
\usepackage{arydshln}
\usepackage{threeparttable}
\newcommand{\myparagraph}[1]{\vspace{0.1em}\noindent\textbf{#1}}
\usepackage{pifont}

\newcommand{\xmark}{\ding{55}}
\newcommand{\ie}{\textit{i}.\textit{e}.}
\newcommand{\eg}{\textit{e}.\textit{g}.}
\begin{document}
\title{Learning Calibrated-Guidance for Object Detection in Aerial Images}
\author{Zongqi~Wei$^{\dag}$,
        Dong Liang$^{\dag}$,
        Dong~Zhang$^{\dag}$,
        Liyan~Zhang$^{*}$,
        Qixiang Geng,
        Mingqiang Wei,~\IEEEmembership{Senior Member,~IEEE}, and
        Huiyu Zhou

\thanks{Z. Wei, D. Liang, L. Zhang, Q. Geng, and M. Wei are with the College of Computer Science and Technology, Nanjing University of Aeronautics and Astronautics, MIIT Key Laboratory of Pattern Analysis and Machine Intelligence, Collaborative Innovation Center of Novel Software Technology and Industrialization, Nanjing 211106, China. 
E-mail: \{weizongqi, liangdong, zhangliyan, gengqixiang, mqwei\}@nuaa.edu.cn.
D. Zhang is with the School of Computer Science and Engineering, Nanjing University of Science and Technology, Nanjing 210094, China. E-mail: dongzhang@njust.edu.cn.
H. Zhou is with the School of Informatics, University of Leicester, Leicester LE1 7RH, United Kingdom. E-mail: hz143@leicester.ac.uk}
\thanks{$\dag$~These authors contributed equally to this work.}
\thanks{$*$~Corresponding author: Liyan Zhang.}
}
\markboth{submission of Journal of Selected Topics in Applied Earth Observations and Remote Sensing}%
{Shell \MakeLowercase{\textit{et al.}}: Bare Demo of IEEEtran.cls for IEEE Journals}
\maketitle
\begin{abstract}
Object detection is one of the most fundamental yet challenging research topics in the domain of computer vision. Recently, the study on this topic in aerial images has made tremendous progress. However, complex background and worse imaging quality are obvious problems in aerial object detection. Most state-of-the-art approaches tend to develop elaborate attention mechanisms for the space-time feature calibrations with arduous computational complexity, while surprisingly ignoring the importance of feature calibrations in channel-wise. In this work, we propose a simple yet effective Calibrated-Guidance (CG) scheme to enhance channel communications in a feature transformer fashion, which can adaptively determine the calibration weights for each channel based on the global feature affinity correlations. Specifically, for a given set of feature maps, CG first computes the feature similarity between each channel and the remaining channels as the intermediary calibration guidance. Then, re-representing each channel by aggregating all the channels weighted together via the guidance operation. Our CG is a general module that can be plugged into any deep neural networks, which is named as CG-Net. To demonstrate its effectiveness and efficiency, extensive experiments are carried out on both oriented object detection task and horizontal object detection task in aerial images. Experimental results on two challenging benchmarks (\ie, DOTA and HRSC2016) demonstrate that our CG-Net can achieve the new state-of-the-art performance in accuracy with a fair computational overhead. The source code has been open sourced at~\href{https://github.com/WeiZongqi/CG-Net}{https://github.com/WeiZongqi/CG-Net}
\end{abstract}

\section{Introduction}
\IEEEPARstart{O}{bject} detection in aerial images is one of the most fundamental yet challenging research tasks, which aims to assign a bounding box with a unique semantic category label to each surficial object in the given aerial images~\cite{ding2019learning, ming2020dynamic, yang2020arbitrary, yang2019scrdet, yang2020dense}. This task is critical for a wide range of downstream tasks, \eg, land resource management, ecological monitoring, and land ecosystem evaluation~\cite{wang2018scene,zhang2019positional}. Thanks to the recent promising development of deep Convolutional Neural Networks (CNNs) in image processing, object detection in aerial images has also made tremendous progress. The state-of-the-art approaches are usually based on a one-stage detector (\eg, RetinaNet~\cite{lin2017focal} and YOLO~\cite{redmon2016you}) or a two-stage detector (\eg, Fast/Faster R-CNN~\cite{girshick2015fast,ren2016faster}) with a CNN as the backbone.
\begin{figure}[t]
\centering
\includegraphics[width=.48\textwidth]{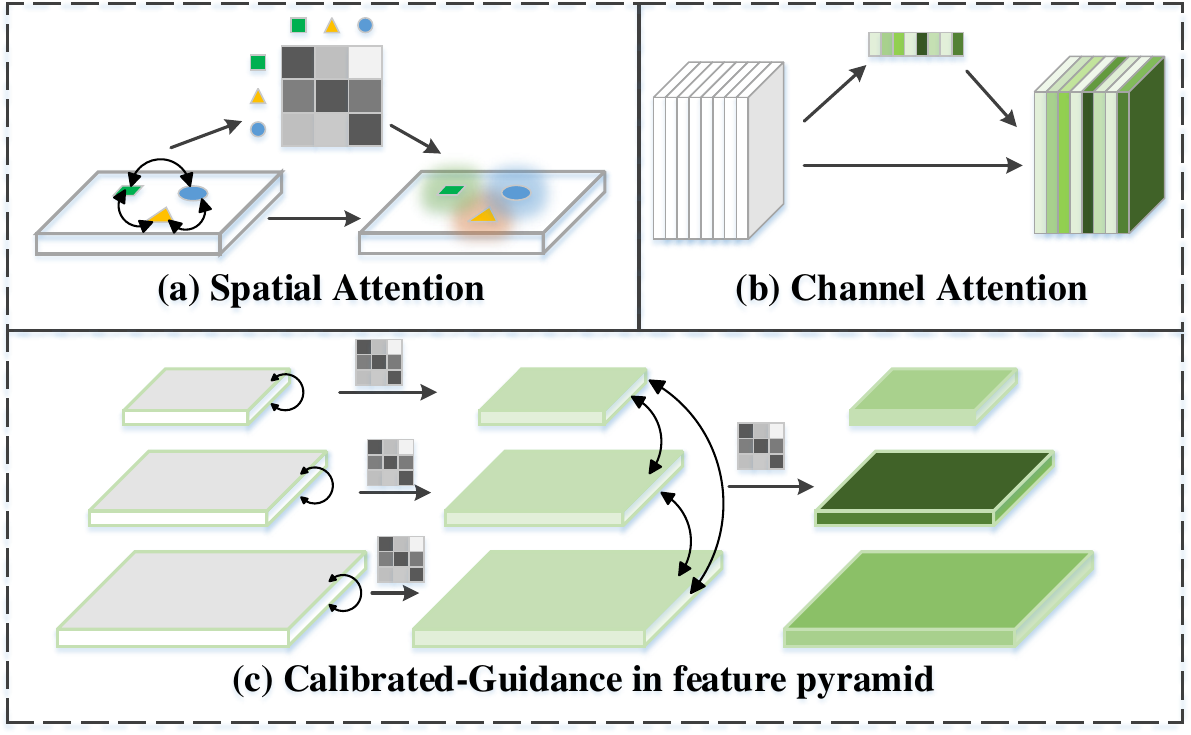}
\caption{Illustrations of the attention-based feature calibration methods, (a) spatial attention mechanism, (b) channel attention mechanism, and (c) our proposed calibrated-guidance.}
\label{fig1}
\end{figure}

\begin{figure*}[t]
\centering
\includegraphics[width=1\textwidth]{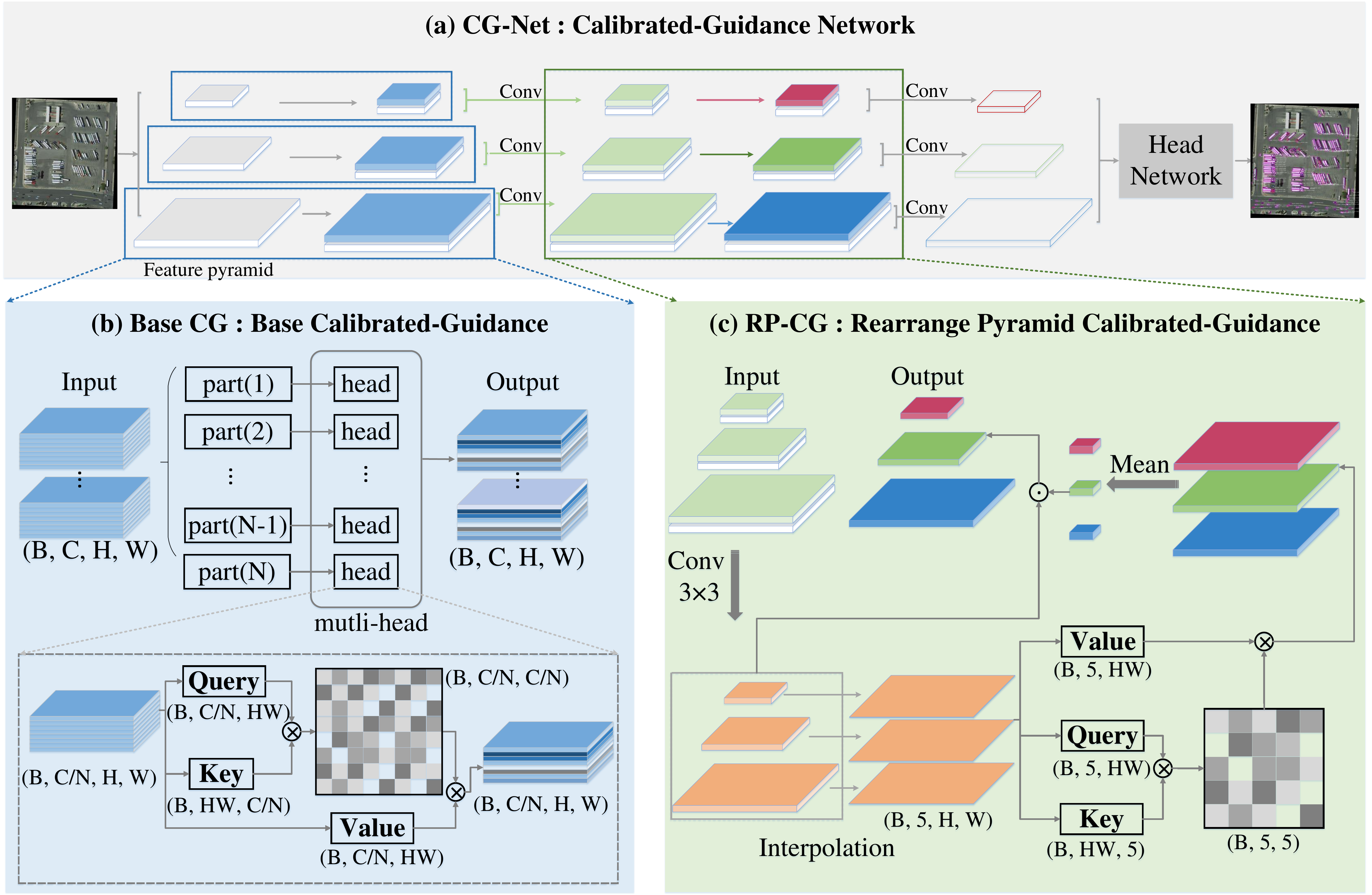}
\caption{Overall architecture of our proposed Calibrated-Guidance Network (CG-Net), where CG  is deployed on both the intra-layer feature maps and the feature pyramid (\ie, the standard feature pyramid network~\cite{lin2017feature}). 
In comparison, feature map with CG module has a stronger representation ability.
After that, we use a task-specific head network for dealing with both oriented and horizontal object detection tasks in aerial images. ResNet~\cite{he2016deep} is used as the backbone network.}
\label{Fig2}
\end{figure*}
Compared to objects in general natural scenes, objects in aerial images usually have smaller size, higher density, objects with different size, worse imaging quality, and more complex background~\cite{xia2018dota,kaminski2021electrical}. Therefore, it is difficult to directly achieve a satisfying recognition performance in aerial images using the existing natural scene object detectors. To this end, state-of-the-art methods focus on developing effective head networks~\cite{ding2019learning}, adaptive dense anchor generators~\cite{ming2020dynamic}, and labeling strategy~\cite{yang2020arbitrary, yang2020dense}. Besides, effective feature learning strategies play a crucial role. 
Because such methods can provide generalized features to improve the model performance.
To this end, a large amount of feature calibration methods based on the attention mechanisms have been proposed to improve the rough feature representations in CNNs~\cite{yang2019scrdet,haut2019remote,chen2019multi,zhang2020causal,wang2018scene,wang2018multiscale}. Conceptually, these attention-based methods can be basically divided into two categories: (\uppercase\expandafter{\romannumeral1}) the spatial-attention-based one, and (\uppercase\expandafter{\romannumeral2}) the channel-attention-based one. For {the first category} (\eg, spatial attention module~\cite{fu2019dual,chen2019multi,wang2018multiscale}, recurrent attention structure~\cite{wang2018scene}, self-attention mechanism~\cite{DETR}, and non-local operation~\cite{wang2018non}), as shown in Figure~\ref{fig1} (a), a global context mapping for each feature position can be obtained by computing the similarities between the feature of each specific  position and all the remaining feature positions~\cite{YuanW18,zhang2021self}. 
Through such an operation, each pixel can obtain the long-range dependence information of the input image. For {the second category} (\eg, channel attention module~\cite{fu2019dual}, and the channel-wise attention~\cite{chen2017sca,chen2019multi}, and the squeeze-and-excitation block~\cite{hu2018squeeze,haut2019remote,yang2019scrdet}), as shown in Figure~\ref{fig1} (b), each channel can obtain a weight that reflects its own importance in object detection, and then integrate the weight into the model by the channel re-weighting manner.

Despite the success of the existing attention-based methods in calibrating features for object detection, we argue that most of these methods on feature calibrations in channels are not enough. That is to say, they cannot introduce channel communications to capture the dependencies between channel feature maps, which has empirically shown their benefits to a wide range of computer visual recognition tasks~\cite{zhang2018shufflenet,wu2018group,howard2019searching,he2020milenas,radosavovic2020designing,yang2017breaking}. Although the existing channel-attention-based methods can enable different channels to obtain different weights, modules (\eg, global average/max pooling) based on their channel feature maps cannot guarantee all the channels have sufficient communications. Therefore, from this point of view, these methods are still local-based.

To address those problem, including different size objects and complex background in aerial images and limitations of existing attention-based methods in calibrating features, in this paper, we propose a simple yet effective Calibrated-Guidance (CG) scheme to enhance channel communications in a feature transformer fashion, which can adaptively determine the calibration weights for each feature channel based on the global feature affinity-pairs. CG is an active feature communication mechanism, as illustrated in Figure~\ref{fig1} (c), which can explicitly introduces feature dependencies in a channel-wise manner. Specifically, CG is applied to the pyramid features, including the inner and inter-layers of the pyramid, and the pyramid layer features are also regarded as the "channel" of overall pyramid features. CG consists of two steps: first, feature similarities (via the dot product operation) between each channel and the remaining channels are computed as the intermediary calibration guidance. Then, we represent each channel by aggregating all the channels weighted together via the guidance. The weighted feature maps has the same spatial size as the input feature maps, but contain richer information about the long-range channel dependency information. For typical problems of aerial images, within and between pyramid layers, we propose Base CG and Rearrange Pyramid CG to realize calibrating features locally and globally.


CG is a general unit that can be plugged into any deep neural network. We name a CNN model deployed  the proposed CG module as CG-Net. The overall architecture is shown in Figure~\ref{Fig2}. To demonstrate its effectiveness and efficiency, we conduct extensive experiments on both oriented object detection task and horizontal object detection task. Experimental results on the challenging benchmarks DOTA~\cite{xia2018dota} and HRSC2016~\cite{lb2017high} for oriented object detection show that our proposed CG-Net can boost substantial improvements compared to the baseline methods and achieves the state-of-the-art performance in accuracy (\ie, $77.89\%$ and $90.58\%$ mAP, respectively) with a fair computational overhead. Besides, experimental results on DOTA~\cite{xia2018dota} for horizontal object detection also validate the flexibility and effectiveness of the proposed CG-Net, which also achieves the new state-of-the-art performance with the accuracy by $78.26\%$ mAP.

In summary, our main contributions are two-fold: 
\begin{itemize}
    \item a simple yet effective CG scheme is proposed to enhance channel communications in a feature transformer fashion, and implements within and between feature pyramid layer to enhance pyramid representation;
    \item  we propose a CG-Net, which can achieve the state-of-the-art oriented and horizontal object detection performance on two challenging benchmarks for aerial images, including DOTA and HRSC2016.
\end{itemize}

\section{Related Work}
\subsection{Object Detection in Aerial Images.}
The purpose of object detection in aerial images is to locate objects of interest on the ground and recognize their categories by a bounding box~\cite{kaminski2021electrical,bouhlel2021abnormal}.
Each bounding box not only contains the object coordinate information, but also contains the category information. Object detection in aerial images can be divided into horizontal-based ones and oriented-based ones. Horizontal object detection aims to detect objects with horizontal bounding boxes~\cite{ren2016faster,dai2016r,lin2017focal,redmon2016you}. Being observed from an overhead perspective, the objects in aerial images present more diversified orientations. Oriented object detection~\cite{liu2017learning,ding2019learning,yang2018r2cnn++,xu2020gliding,qian2019learning,ming2021cfc,yang2019scrdet,jiang2017r2cnn,yang2019r3det,ming2020dynamic,yang2020arbitrary,wei2020oriented,li2020novel,yang2020dense} is an extension of horizontal object detection to accurately outline the objects, especially those with large aspect ratios. 

Based on horizontal object detection, rotating boxes are important learning parts in oriented object detection. There are many methods on how to rotate boxes. CSL~\cite{yang2020arbitrary} design a detection frame by transforming angular prediction form a regression to a classification task. Gliding Vertex~\cite{xu2020gliding} glides the vertex of the horizontal bounding box (regressing four length ratios characterizing the relative gliding offset on each corresponding side) on each corresponding side to accurately describe a multi-oriented object.
DAL~\cite{ming2020dynamic} propose a dynamic anchor learning method, which utilizes the newly defined matching degree to comprehensively evaluate the localization potential of the anchors. RoI Trans~\cite{ding2019learning} proposes a ROI Transformer to address the mismatches between the Region of Interests (RoIs) and objects on training. CFC-Net~\cite{ming2021cfc} proposes a Critical Feature Capturing Network to address problems of discriminative features in object detection in refining preset anchors, building powerful feature representation and optimizing label assignment. R-RPN~\cite{li2020novel} overcomes the limitation of ROI pooling when extracting ships features with various aspect ratios. For fast and accurate oriented object detection, R$^3$Det~\cite{yang2019r3det} and O$^2$-DNet~\cite{wei2020oriented} make attempts in one-stage model with RetinaNet and anchor free structures. Based on R$^3$Det, R$^3$Det-DCL~\cite{yang2020dense} designs Densely Coded Labels~(DCL) for angle classification, which replaces the Sparsely Coded Label (SCL) in classification-based detectors before, and reduces three times training speed, further bringing notable improvements in accuracy of detection tasks. What's more, for oriented object detection, SCRDet~\cite{yang2019scrdet} combines pixel and channel attention network for small and cluttered objects. DEA~\cite{liang2021anchor} leverages a sample discriminator to realize interactive sample screening between an anchor-based unit and an anchor-free unit to generate eligible samples in aerial images detection.

From the presentation form of bounding boxes, oriented object detection can be more suitable for aerial object detection, because it contains the orientation information of objects with more accurate bounding-box. In this work, we consider both oriented and horizontal aerial object detection tasks and develop a pipeline line to benefit both of them.

\subsection{Feature Calibration over Images.}
The purpose of feature calibration is to refine feature maps through the existing information, so as to further improve the representation ability. Currently, most of the state-of-the-art methods are designed from the perspective of feature calibration to deal with the challenges of complex background and noise in object detection~\cite{fu2019dual,wang2018non,hu2018squeeze,chen2019multi,wang2018multiscale}. Among those methods, attention-based ones are proposed to calibrate features from two aspects, including spatial-attention and channel-attention-based. 

Spatial-attention-based mechanisms capture object positions in the spatial dimension. Position attention module~\cite{fu2019dual}/Non-local operation~\cite{wang2018non} build rich contexts on local features by using a self-attention mechanism. Transformer~\cite{vaswani2017attention} is the first sequence transduction model combined with multi-headed self-attention. DETR~\cite{DETR} is proposed to explore the relationship between objects in the global context, which is of precision similar to those of the two-stage detectors, but has a weakness on detecting large objects with high computational overheads~\cite{yan2018participation,yan2020higcin}. In aerial image  analysis, 
ARCNet~\cite{wang2018scene} utilizes a recurrent attention structure to squeeze high-level semantic features for learning to reduce parameters.
Channel-attention-based mechanisms allocate resources for channels referring to their importance.  SENet~\cite{hu2018squeeze} utilizes a squeeze-and-excitation block to implement dynamic channel-wise feature re-calibration. For obtaining better feature representations, DANet~\cite{fu2019dual} utilizes a channel attention module to capture contextual relationships based on the self-attention mechanism. In aerial image, a residual-based network combining channel attention~\cite{haut2019remote} is used to learn the most relevant high-frequency features.

There are also some works that combine spatial attention with channel-wise attention together, \eg, SCA-CNN~\cite{chen2017sca}, and DONet~\cite{zhang2019cascaded}. 
These methods take advantage of both channel-wise attention and spatial-wise attention.
Besides, to address considerable interference of complex background in aerial detection, multi-scale spatial and channel-wise attention mechanism~\cite{chen2019multi} is proposed to strengthen the object region in aerial detection task.
Despite the success of the existing attention-based methods, they are not sufficient for feature calibration in channels. In this work, we propose a simple yet effective CG scheme to enhance channel communications in a feature transformer fashion, which can adaptively determine the calibration weights for each channel based on the global feature affinity-pairs.

\section{Methodology}
In this section, we show the technical details of our proposed Calibrated-Guidance Network (CG-Net) for object detection in aerial images. Specifically, we first revisit the channel attention mechanism on images in Section~\ref{sec3.1}. Then, our proposed Calibrated-Guidance (CG) module which can enhance channel communications is described in Section~\ref{sec3.2}. After that, we introduce how to implement CG on the base CNNs' feature maps (\ie, Base CG) and on an intra-network feature pyramid (\ie, Rearranged Pyramid CG) for object detection in aerial images in Section~\ref{sec3.3} and Section~\ref{sec3.4}. Finally, we show the details of the network architecture in Section~\ref{sec3.5}. 
\subsection{Channel Attention Revisited}\label{sec3.1}
Channel-wise Attention (CA) module utilizes the inter-dependencies between the channels to emphasize the important ones by weighting the similarity matrix. To be specific, CA operates on queries~(\textbf{Q}), keys~(\textbf{K}) and values~(\textbf{V}) among a set of single-scale feature maps $\textbf{X}$, and the improved version $\textbf{X}'$ has the same scale as the original $\textbf{X}$. For a given set of feature maps $\textbf{X} \in \mathbb{R}^{W \times H \times C}$, where $W$, $H$ and $C$ are width, height and channel dimension, respectively, CA implementation can be formulated as:
\begin{equation}
\begin{aligned}
\textbf{\textrm{Input}}&: \textbf{q}_i, \textbf{k}_j, \textbf{v}_j \\
\textbf{\textrm{Similarity}}&: \textbf{s}_{i,j} = F_{\textrm{sim}}(\textbf{q}_{i}, \textbf{k}_{j}) \\
\textbf{\textrm{Weight}}&: \textbf{w}_{i,j} = F_{\textrm{nom}}(\textbf{s}_{i,j}) \\
\textbf{\textrm{Output}}&: \textbf{X}_i' = \sum_{j}F_{\textrm{mul}}(\textbf{w}_{i,j},\textbf{v}_j),
\label{eq1}
\end{aligned}
\end{equation}
where $\textbf{q}_i = f_q(\textbf{X}_i) \in \textbf{Q} $ is the $i^{th}$ query; $\textbf{k}_j = f_k(\textbf{X}_j) \in \textbf{K} $ and $\textbf{v}_j = f_v(\textbf{X}_j) \in \textbf{V} $ are the $j^{th}$ key/value pair; $f_q(\cdot)$, $f_k(\cdot)$ and $f_v(\cdot)$ denote the query/key/value channel transformer functions~\cite{vaswani2017attention, DETR}; $\textbf{X}_i$ and $\textbf{X}_j$ denote the $i^{th}$ and $j^{th}$ channel feature in $\textbf{X}$; $F_\textrm{sim}$ is the dot product similarity function; $F_{\textrm{nom}}$ is the softmax normalization function; $F_{\textrm{mul}}$ denotes matrix dot multiplication; $\textbf{X}'_i$ is the $i^{th}$ channel feature in the transformed feature map $\textbf{X}'$, and the response of $i^{th}$ channel feature is computed by $j^{th}$ ones that enumerates all possible channels. Although CA can enable different channels to obtain different weights, the coarse operation based on the entire channel feature maps (\ie, without the grouped feature representations~\cite{zhang2018shufflenet,vaswani2017attention,radosavovic2020designing,yang2017breaking}) cannot enable all the channels to have sufficient communications, which has been empirically shown its importance in a large range of computer vision tasks. As a result, the ability to feature representation is limited.
\subsection{Calibrated-Guidance~(CG)}
\label{sec3.2}
We propose CG to enhance feature channel communications in a feature transformer fashion, which can adaptively determine the calibration weights for the channels based on the global feature affinity-pairs. Its detailed structure is illustrated in Figure~\ref{Fig2}. CG is inspired by the transformer mechanism and the difference is that we combine the multi-head representations, and concatenate the original feature maps and the calibrated features, then use a convolution layer to produce the enhanced feature maps as output. 


We deploy the multi-head architecture to focus on richer channel feature representations. Multi-head in ViT~\cite{dosovitskiy2020image} and DETR~\cite{DETR} can provide more feature selection when extracting features. Multi-head structure complements features by learning different contents, which is more sufficient than one head. Analysis work~\cite{voita2019analyzing} finds that important ones in multi-head have one or more specialized and interpretable functions in the model, which indirectly shows the necessity of adopting multi-head structure.

First, we divide query and key into $N$ parts in the channel dimension. Then, we feed the divided feature with shape~($B, C/N, H, W$) into each head, where each structure is a CG module ($B$ is batch size). For $n^{th}$ head module, the shape of similarity matrix $\textbf{s}^n$ is ($B, C/N, C/N$), which can be expressed as:
\begin{equation}
\begin{aligned}
\textbf{s}^n = 
\begin{bmatrix}
	  w^{nC/N,nC/N}  & \cdots   & w^{(n+1)C/N,0}  \\
	\vdots   & \ddots   & \vdots  \\
	w^{0,(n+1)C/N}  & \cdots\  & w^{(n+1)C/N,(n+1)C/N}  \\
\end{bmatrix},
\label{eq(s)}
\end{aligned}
\end{equation}
where each $w$ denotes the learnable similarity scalar. After that, the outputs of these head modules (\ie, the partial result) are concatenated together to produce the holistic output feature maps, which have the same shape as the original feature maps. The above process can be formulated as:
\begin{equation}
\begin{aligned}
\textbf{\textrm{Weight}}&: \textbf{w}_{i,j}^n = F_{\textrm{nom}}(\textbf{s}_{i,j}^n) \\
\textbf{\textrm{Partial Result}}&: \textbf{X}_{i}^{n} = \sum_{j}F_{\textrm{mul}}(\textbf{w}^n_{i,j},\textbf{v}_{j,n})\\
\textbf{\textrm{Holistic Output}}&: \textbf{X}' = F_{\textrm{con}}(\textbf{X}_{i}^{n}),
\label{eq2}
\end{aligned}
\end{equation}
where $\textbf{s}_{i,j}^n$ and $\textbf{w}_{i,j}^n$ denote the $n^{th}$ partial similarity weight of the $i^{th}$ and $j^{th}$ channel features and the normalized one. The $i^{th}$ channel feature is calculated by other channel features. $\textbf{v}_{j,n}$ denotes the $j^{th}$ value of the $n^{th}$ head. $F_{\textrm{con}}$ is used for feature concatenation in the channel dimension. Compared to the previous transformer-based approaches, the multi-head CG has lower computational complexity, $\textit{O}(NC^2)$  both in time and space, while the previous ones have the computational complexity of $\textit{O}(NH^2W^2)$. Compared to CA, our proposed CG implements on pyramid features have the following three advantages:
(\romannumeral1) CG is designed for the enhancement of communications within and between feature pyramid layers, while most of the previous ones are used to capture the long-range dependencies in space and channel within features. (\romannumeral2) CG is based on the multi-head structure, which has its unique tendency of feature representation in different feature spaces~\cite{vaswani2017attention,bello2019attention}. Hence CG can provide an enhanced feature representation. (\romannumeral3) CG is designed for object detection in aerial images. By enhancing feature pyramid representation, CG can solve complex background and worse imaging quality problems in aerial images, then obtain a more accurate proposals in head network~(in Section~\ref{sec4.2}). Experimental results (Section~\ref{sec4.3}) show that CG can improve the state-of-the-art performance swimmingly on both oriented and horizontal tasks. Two CG implements of Base CG and Rearranged Pyramid CG show as follows.
\subsection{Base CG}\label{sec3.3}
Given an arbitrary aerial image, we can extract a set of feature maps by a fully convolution network. For these feature maps, CG can directly achieve calibrated-guidance practice to enhance channel communications and adaptively determine the calibration weight for each channel. Its detailed architecture in a level of the feature pyramid (\ie, feature maps with the same scale) is illustrated in Figure~\ref{Fig2} (b). Since this CG implementation is performed on the basic feature maps, we call it Base CG. 
Base CG is a general unit, which works on the backbone network.

Compared to other existing head-network-based task-specific methods~\cite{yang2019scrdet,lin2019ienet}, it is more universal and can facilitate a wide range of downstream recognition tasks. Our Base CG improves feature extraction, and the results can be seen from the ablation experiments shown in Section~\ref{sec4.2}.

\subsection{Rearranged Pyramid CG}\label{sec3.4}
Feature pyramid has shown its effectiveness in a wide range of computer vision tasks~\cite{lin2017feature,lin2017focal,zhang2020feature}. In this section, we show how to implement our Calibrated-Guidance on a feature pyramid (\ie, the proposed Rearranged Pyramid CG~(RP-CG)). Compared to the existing feature calibration methods on the in-network feature pyramid~\cite{pang2019libra,ghiasi2019fpn,qin2019thundernet}, our RP-CG has lower computational complexity and fewer model parameters (details are shown in Section~\ref{sec4.1}). The RP-CG module works on an extracted feature pyramid from the feature pyramid network~\cite{lin2017feature}, whose architecture is illustrated in Figure~\ref{Fig2} (c). 

From the perspective of levels inside the feature pyramid, each level can been seen as local features, \ie, only part of the features of the input image are captured. In order to emphasize the most suitable feature in the channel dimension of the feature pyramid, combining global and local information is crucial in feature extraction. In our work, RP-CG focuses on weighting different features among pyramid levels $\textbf{X}_{P2-P6}$ following work~\cite{lin2017feature, zhang2020feature}.
As illustrated in Figure~\ref{Fig2} (c), we apply CG between 5 levels of the feature pyramid to fully communicate levels’ information. In our implementation, firstly, we reduce the channel dimension and launch interpolation on pyramid features ${\textbf{X}}_{P2-P6}$ to generate the same scale features (same scale as the largest one: $P2$) and then concatenate them as $\overline{\textbf{X}}_{P2-P6}$, which is expressed as: 
\begin{equation}
\begin{aligned}
\overline{\textbf{X}}_{P2-P6} = F_{\textrm{intp}}(\textbf{X}_{P2-P6}),
\label{eq3}
\end{aligned}
\end{equation}
where $F_{\textrm{intp}}$ is a channel dimension reduction and scale interpolation function. The shape of output feature $\overline{\textbf{X}}_{P2-P6}$ is ($B, 5, H_{p2}, W_{p2}$). Then, same as Base CG, RP-CG produces the output $\overline{\textbf{X}}_i'$ from input $\textbf{q}_i, \textbf{k}_j$ and $\textbf{v}_j$ by learning the weight between the query and the key. The interaction is formulated as:
\begin{equation}
\begin{aligned}
\textbf{\textrm{Input}} & : \textbf{X}_{P2-P6} \\
\textbf{\textrm{Interpolation}} & : \overline{\textbf{X}}_{P2-P6} \\
 \textbf{\textrm{Extraction}} & : \textbf{q}_i, \textbf{k}_j, \textbf{v}_j \\
\textbf{\textrm{Similarity}} & : \textbf{s}_{i,j} = F_\textrm{sim}(\textbf{q}_i, \textbf{k}_j) \\
\textbf{\textrm{Weight}} & :\textbf{w}_{i,j} = F_{\textrm{nom}}(\textbf{s}_{i,j}) \\
\textbf{\textrm{Output}} & : \overline{\textbf{X}}_i' = \sum_{j}F_{\textrm{mul}}(\textbf{w}_{i,j}, \textbf{v}_j)\\
\textbf{\textrm{Holistic Output}}&: \overline{\textbf{X}}^{rpcg}_{P2-P6} = F_{\textrm{con}}(\overline{\textbf{X}}_i'),
\label{eq5}
\end{aligned}
\end{equation}
where $\overline{\textbf{X}}_i'$ is the $i^{th}$ level feature in transformed feature map $\overline{\textbf{X}}^{rpcg}_{P2-P6}$ with shape ($B, 5, H_{p2}, W_{p2}$). $\overline{\textbf{X}}^{rpcg}_{P2-P6}$ realizes global channel communication in pyramid features, but we need to find the right way to feed back to pyramid features. 

In addition, there have been multitudes of methods to verify the effectiveness of the combination of global and local information in visual recognition, and our method is global in essence. To this end, combining our RP-CG with the existing local channel attention method is a natural choice. In this work, the classical channel attention~\cite{hu2018squeeze} is chosen. Based on this, the overall structure of our proposed Rearrange Pyramid Calibrated-Guidance module can be expressed as:
\begin{equation}
\begin{aligned}
\textbf{\textrm{Weight}} & :\overline{\textbf{X}}^{\textrm{mean}(rpcg)}_{P2-P6} = F_{mean}(\overline{\textbf{X}}^{rpcg}_{P2-P6})\\
\textbf{\textrm{Scale}} & : \overline{\textbf{X}}_{P2-P6}' = \overline{\textbf{X}}^{mean(rpcg)}_{P2-P6} \otimes \textbf{X}_{P2-P6}\\
\textbf{\textrm{Output}} & : \overline{\textbf{X}}_{P2-P6}^{final} = F_{conv}(\overline{\textbf{X}}_{P2-P6}' \oplus \textbf{X}_{P2-P6}).
\label{eq6}
\end{aligned}
\end{equation}
The output from $\overline{\textbf{X}}_{P2-P6}$ are divided into $5$ parts~($P2-P6$). $\overline{\textbf{X}}^{rpcg}_{P2-P6}$ is the overall feature after we have weighted $\overline{\textbf{X}}_{P2-P6}$. We use $F_{\textrm{mean}}$ to derive the weighting parameter to distinguish different scales' features, and it includes the operation of using the mean value as the weighting parameter for each pyramid's levels, which is then resized to the same scale of the original level feature.
$\otimes$ is matrix cross multiplication, and $\oplus$ is channel concatenation. $\overline{\textbf{X}}_{P2-P6}'$ is the calibrated feature with the same size as the original feature pyramid. We get final output $\overline{\textbf{X}}_{P2-P6}^{final}$ from convolution $F_{conv}$, which is to reduce the channel to the original size.


\subsection{Network Architecture}
\label{sec3.5}
CG can help the model learn richer communication information between feature channels, so it is suitable for object detection task in aerial images. In this paper, we build a Calibrated-Guidance network~(CG-Net) for both oriented and horizontal object detection tasks of aerial images. The overall architecture is illustrated in Figure~\ref{Fig2}. CG-Net is based on our proposed \textbf{Base CG} (in Figure~\ref{Fig2} (b)) and \textbf{RP-CG} (in Figure~\ref{Fig2} (c)) for transforming pyramid features. Specifically, we deploy ResNet~\cite{he2016deep} as backbone following~\cite{ding2019learning}, which has been pre-trained on the ImageNet~\cite{deng2009imagenet}. Then, we produce a feature pyramid from the feature pyramid network~\cite{lin2017feature}. For this feature pyramid, we firstly apply \textbf{Base CG} in the feature maps from each level of the pyramid. After that, we deploy the \textbf{RP-CG} to produce a new feature pyramid that realizes global and local communication in the feature pyramid. Then, we concatenate the original feature maps with the calibrated ones together in the channel dimension and reduce the dimensionality of the concatenated feature maps into $256$ channels by a $3$ $\times$ $3$ convolution. Finally, we use the head network from the RoI transformer~\cite{ding2019learning} for oriented object detection and a standard Faster R-CNN~\cite{ren2016faster} for horizontal object detection.

\section{Experiments}
To demonstrate the effectiveness and efficiency of our proposed method, experiments are carried out on both oriented object detection task and horizontal object detection tasks in aerial images. In what follows, we first show experiments settings including datasets, image size, baseline model, hyper-parameters, implementation details and evaluation metrics in Section~\ref{sec4.1}. Then we show some ablation results including some quantitative and qualitative experimental results in Section~\ref{sec4.2}. Finally, we show result comparisons with state-of-the-art methods in Section~\ref{sec4.3}.
\subsection{Experimental Setup}\label{sec4.1}
In our work, two challenging datasets are selected in experiments, which are A Large-Scale Dataset for Object Detection in Aerial Images (DOTA) dataset~\cite{xia2018dota} and High Resolution Ship Collections 2016 (HRSC2016) dataset~\cite{lb2017high}.  DOTA is used for both oriented and horizontal object detection. HRSC2016 is used for only oriented object detection.
\begin{itemize}
\item \emph{\textbf{DOTA}}~\cite{xia2018dota} is the largest aerial images datasets for object detection and includes oriented and horizontal bounding boxes. From different platforms and sensors, DOTA contains $188,282$ annotated instances in $2,806$ aerial images, and have $15$ common object categories, like Plane (PL), Bridge (BR), Baseball diamond (BD), Ground track field (GTF), Large vehicle (LV), Small vehicle (SV), Ship (SH), Basketball court (BC), Tennis court (TC), Storage tank (ST), Roundabout (RA), Soccer-ball field (SBF), Harbor (HA), Swimming pool (SP), and Helicopter (HC). Images range in size between about $800 \times 800$ and $4,000 \times 4,000$ pixels and contain objects rendered in various scales, orientations, and shapes. For dataset split, we follow the setting of work~\cite{xia2018dota,yang2019scrdet}, and randomly select $1/2$ of the original images as the training set, $1/3$ as the testing set, and $1/6$ as the validation set.
\item \emph{\textbf{HRSC2016}}~\cite{lb2017high} is a ship detection dataset of aerial images with challenging problems like arbitrary orientations and large aspect ratios. HRSC contains $20$ ship categories with various appearances in $1061$ images, collected from 6 harbors by Google Earth. Images range in size between about $300 \times 300$ and $1500 \times 900$ pixels. For dataset split, we follow the setting of work~\cite{lb2017high}, and the ratio of the training, validation, and test sets is $5:2:5$, respectively including $436$ images, $181$ images, and $444$ images. 
\end{itemize}
\begin{figure}[t]
\centering
\includegraphics[width=.5\textwidth,trim=300 100 340 190,clip]{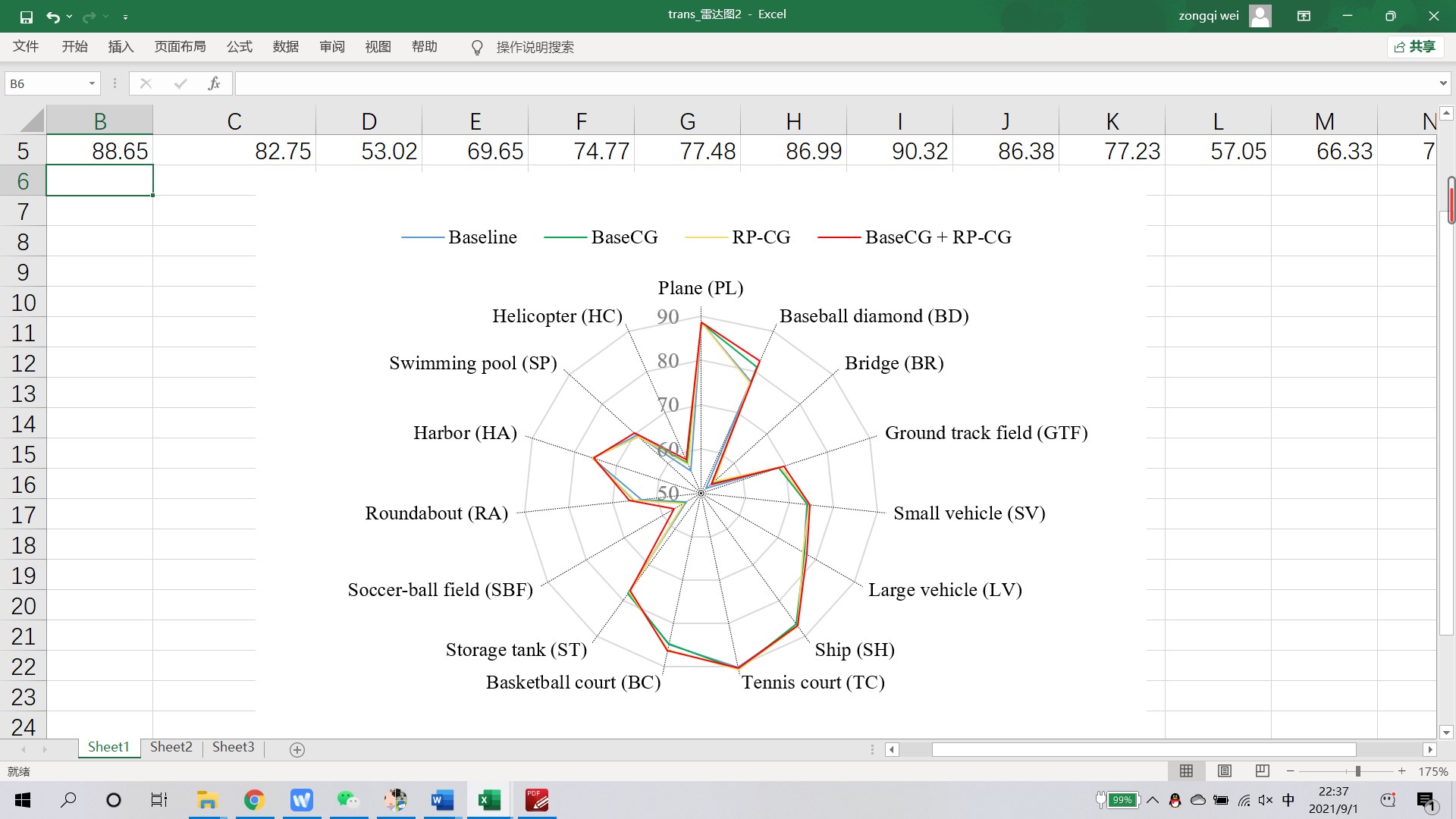}
\caption{Radar chart for each category of object in DOTA~\cite{xia2018dota} dataset.  Different colored lines represent different detectors. The larger areas enclosed by the outer line, the better recognition performance of the corresponding method. The value in this figure denotes the mAP.}
\label{Figure_radar1}
\end{figure}

\begin{figure} 
\centering
\includegraphics[width=.35\textwidth]{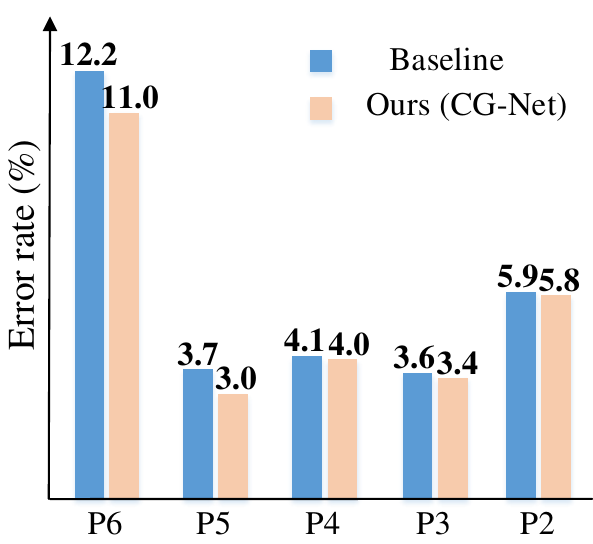}
\caption{Mismatching error rate comparison between the baseline and the proposed method on DOTA~\cite{xia2018dota} with ResNet-101~\cite{he2016deep} for oriented object detection. The lower the better.}
\label{Figure_histogram}
\end{figure}
\begin{figure*}[t]
\centering
\includegraphics[width=.9\textwidth]{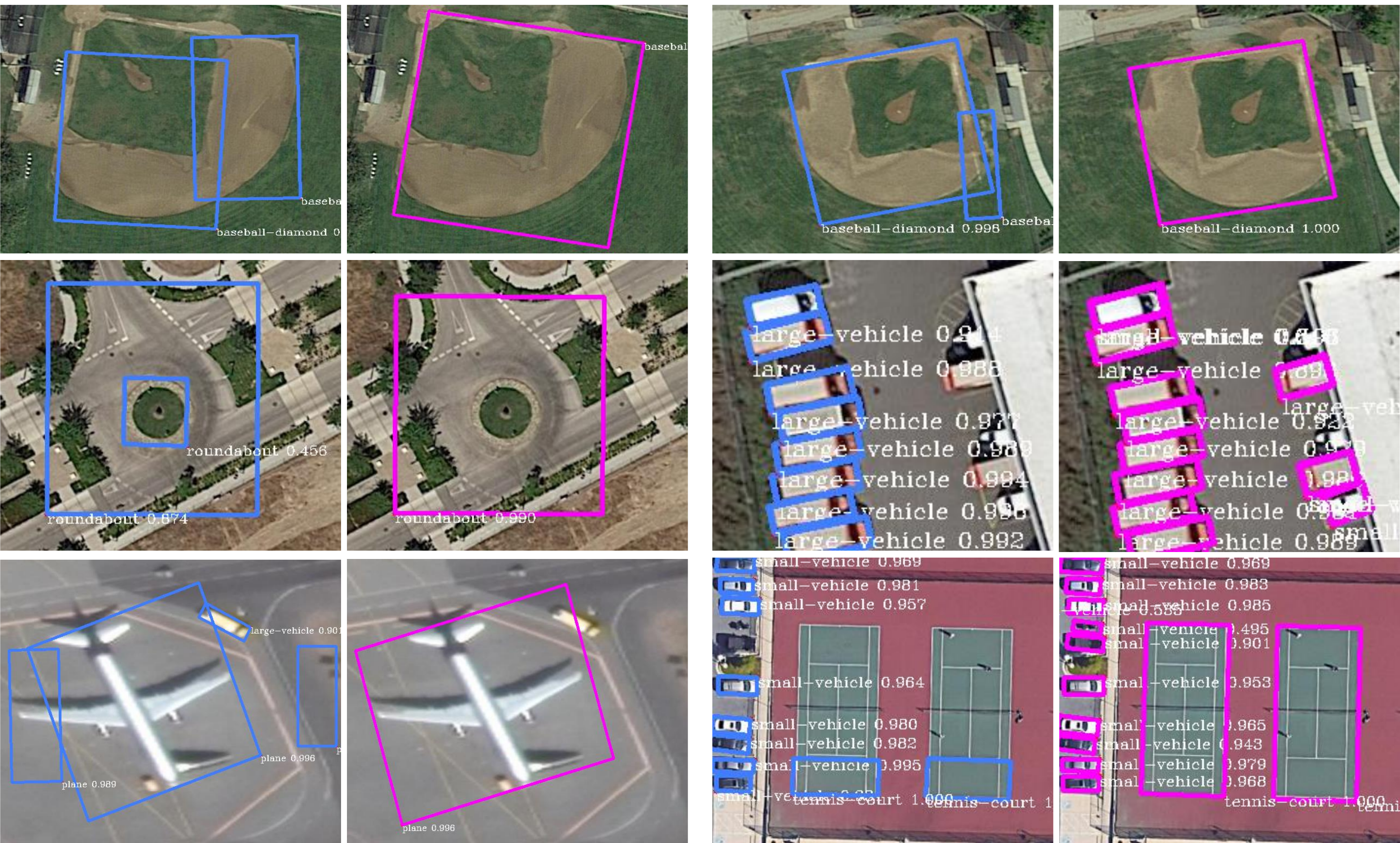}
\caption{Comparison to the baseline on DOTA~\cite{xia2018dota} for oriented object detection with ResNet-101~\cite{he2016deep}. The figures with ${\color{blue}{\textrm{blue~boxes}}}$ are the results of the baseline and ${\color{magenta}{\textrm{pink~boxes}}}$ are the results of our proposed CG-Net.}
\label{Figure_view_new}
\end{figure*}
Due to inconsistent image sizes in the experimental datasets and taking into account the training efficiency and effect for DOTA and HRSC2016, we follow benchmark~\cite{ding2019learning} setting and generate a list of $1,024 \times 1,024$ patches based on original images using $824$ stride for training, validation and test sets.
\begin{table}[t]	
\begin{center}
\renewcommand\arraystretch{1.5} 
\setlength{\tabcolsep}{3pt}{ 
			\begin{tabular}{cc|cc|c}
				\hline
				\hline
				Backbone & +Ours & GFLOPs / FPS & $\#$Params (M) & mAP (\%) \\
				\hline
				\multirow{2}*{ResNet-50} & \xmark  & 211.30 / 25.0 & 41.20 & 73.26 \\
				~ & 	\checkmark & 366.32 / 14.8 & 42.99 & 74.21$_{\color{red}{ +0.95 }}$ \\
				\hline
				\multirow{2}*{ResNet-101} &  \xmark	& 289.26 / 20.9 & 60.19 & 73.06 \\
				~ & 	\checkmark& 444.21 / 13.4 & 61.98 & \textbf{74.30}$_{\color{red}{ +1.24 }}$ \\
				\hline
				\multirow{2}*{ResNet-152} &  \xmark	& 367.23 / 17.0 & 75.83 & 72.78 \\
				~ & 	\checkmark& 522.19 / 12.1 & 77.63 & 73.53$_{\color{red}{ +0.75 }}$ \\
				\hline
				\hline
			\end{tabular}}
	\end{center}
	\caption{The effectiveness of our proposed methods with different backbone networks on the test set of DOTA~\cite{xia2018dota} for oriented object detection. ``+ Ours'' indicates the implementation of our proposed Base CG and RP-CG units on the backbone networks.}
	\label{t_param_flops}
\end{table}
\begin{table}[t]	
\begin{center}
\renewcommand\arraystretch{1.5} 
\setlength{\tabcolsep}{3pt}{ 
			\begin{tabular}{ccc|cc|c}
				\hline
				\hline
				Baseline & CG & Multi-head & GFLOPs / FPS & $\#$Params (M) & mAP (\%) \\
				\hline
				\checkmark & \checkmark & \xmark	& 461.49 / 11.9 & 62.34 & 73.62 \\
				\hline
				\checkmark & \checkmark & \checkmark& 444.21 / 13.4 & 61.98 & \textbf{74.30}$_{\color{red}{ +0.68 }}$ \\
				\hline
				\hline
			\end{tabular}}
	\end{center}
	\caption{The effectiveness of Multi-head structure on the test set of DOTA~\cite{xia2018dota} for oriented object detection. ``CG'' indicates the implementation of our proposed CG units on the backbone networks. ``Multi-head'' means combine Multi-head structure in CG blocks.}
	\label{t_multihead}
\end{table}

Our baseline model is Faster R-CNN~\cite{ren2016faster}, which is the standard two-stage detector in object detection and backbone utilizes ResNet-101. We adopt FPN~\cite{lin2017feature} as neck network to construct a feature pyramid with predefined anchors on pyramid level P2 - P6. In oriented object detection, we utilize RoI-Transformer~\cite{ding2019learning} as the rotated head network that transforms horizontal proposals into rotated ones. For comparison fairly, all parameter and experimental settings are strictly consistent as those reported in~\cite{ding2019learning, xia2018dota, lb2017high}. The entire network is trained by end-to-end style without any extra rotation setting.

Although experience shows that the adjustment of hyperparameters is conducive to the further improvement of model performance, it is necessary for the fairness of comparison. In this paper, following~\cite{ding2019learning, ming2020dynamic}, For DOTA and HRSC2016, anchor size is set to \{$8^2$\} with \{$1/2$, $1$, $2$\} aspect ratios and \{$4$, $8$, $16$, $32$, $64$\} anchor strides of each pyramid level in horizontal anchors. To compare fairly and verify the effectiveness of the proposed method, we conducted ablation studies based on DOTA, and we avoid combining any other data augmentation or bells-and-whistles training strategy. When comparing with SOTA methods on DOTA and HRSC2016, like~\cite{ding2019learning, yang2019scrdet, ming2020dynamic}, we only add an augmentation with random rotation from the angles of ($0, 90, 180,270$).
For multi-head, $N$ can be seen as a hyperparameter used to divide channels and set the number of multi-heads in Base CG. The dividing feature can provide more feature selection for model learning, and if $N$ is large, it will weaken the communication ability among channel. Following parameter setting of previous work~\cite{zhang2020feature} and parameter adjustment, we set $N$ to 2 in our final network.

In our work, the learning rate is $0.005$ initially and conducts $0.0001$ weight decay and $0.9$ momentum in the SGD optimizer. Training iterations are set to 80k and 20k for DOTA and HRSC2016 following~\cite{xia2018dota, lb2017high}. In the testing step, we do not use any testing augmentation, such as multi-scale input or TTA. Besides above, experiments are conducted on two RTX2080Ti.


For evaluation, the results can be obtained from DOTA official evaluation server \footnote{https://captain-whu.github.io/DOTA/} by submitting predictions files. The mean Average Precision~(mAP) of each category and entire is used to evaluate the model and analyze the result distribution following~\cite{xia2018dota}. What's more, GFLOPs / FPS and model Parameters ($\#$Params) are adopted to verify efficiency in the model, which is used to evaluate the computational complexity and runtime efficiency of the model. 


\begin{figure*}[t]
	\centering
	\includegraphics[width=1\textwidth]{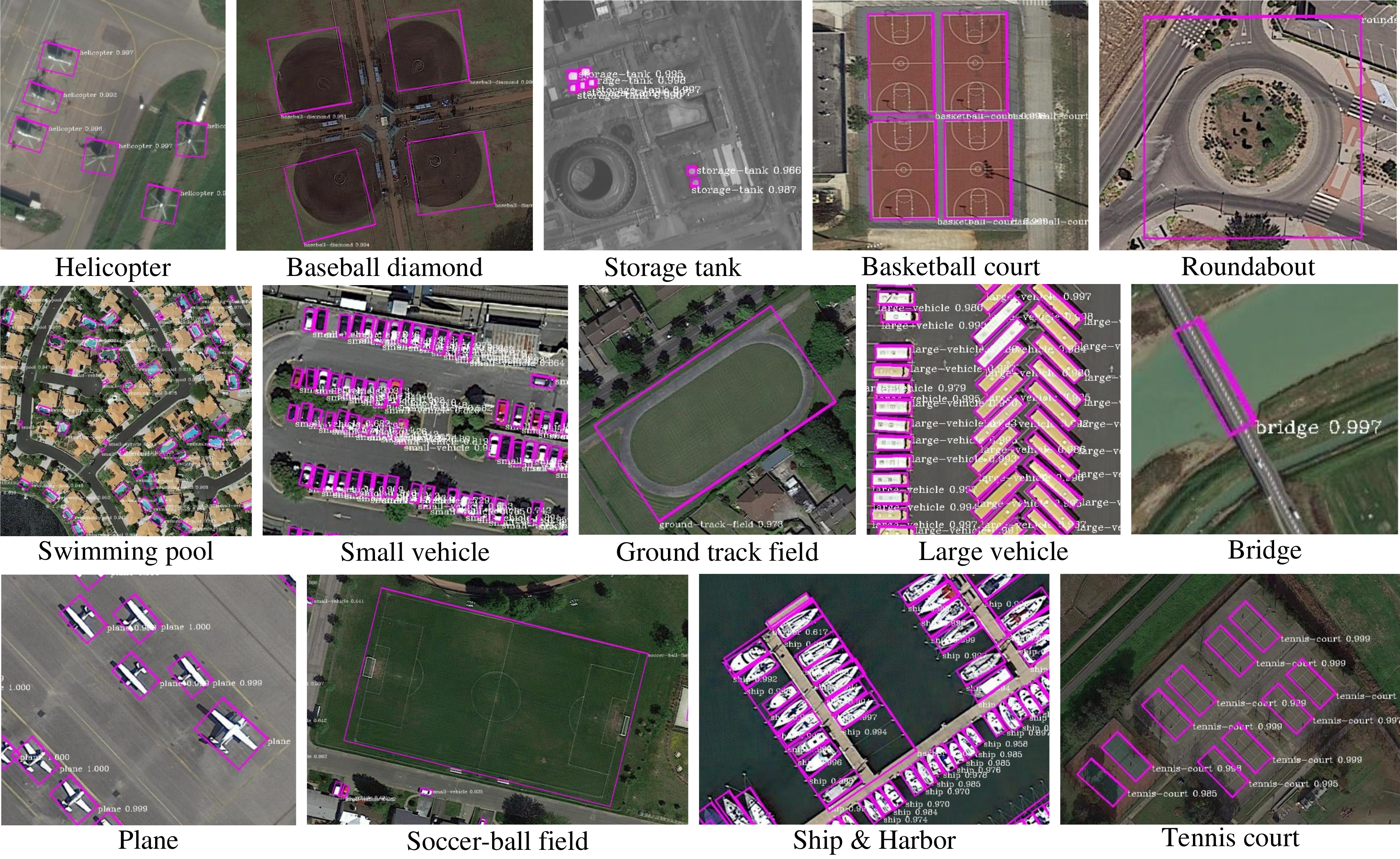}
	\caption{Visualization results for oriented object detection on the test set of DOTA~\cite{xia2018dota}}
	\label{Figure_view_all}
\end{figure*}

\subsection{Ablation Study}\label{sec4.2}
Based on DOTA~\cite{xia2018dota}, we carry out ablations study for oriented object detection in aerial images, which is aimed to: \textbf{(1)} verify the efficiency and effectiveness of different backbone networks combining our proposed methods; \textbf{(2)} verify the effectiveness of the two proposed units on base CNN feature maps (\ie, Base CG) and a feature pyramid (\ie, RP-CG); \textbf{(3)} compare different attention structure with our proposed methods; \textbf{(4)} explore the improvements of RPN input for aerial object detection; \textbf{(5)} reveal mismatching error rates on different scales; and \textbf{(6)} show some visual comparisons. The details are as follows:

\myparagraph{(1) Different backbones} 

In Table~\ref{t_param_flops}, the experimental results show different backbone networks results on the test set of DOTA, containing ResNet-50, ResNet-101, and ResNet-152. We contrast GFLOPs/FPS/$\#$Params/mAP and improvements from the combination of our module. We can observe that combining our units to the backbone can increase mAP by $0.95\%$, $1.24\%$, and $0.75\%$.
Besides, $\#$Params and GFLOPs / FPS are reported for comparisons of model efficiency. Using Base CG and RP-CG increases computational costs; for example, it brings an average of $1.80$ M model $\#$Params with around $155$ GFLOPs increment, and with around 5-10 FPS reduction on these three backbones. Considering the mean Average Precision and Computational complexity, ResNet-101 is selected as our backbone network in experiments. 

\myparagraph{(2) The proposed units} 

In Table~\ref{t1}, we show our proposed units and their combined performance on ResNet-101. We can observe that Base CG and RP-CG respectively bring $0.58\%$ and $0.46\%$ improvements for the bounding box mAP. The corresponding each-category mAP radar chart for oriented object detection is in Figure~\ref{Figure_radar1}, to show the trend of the performance change. 
Combining Base CG and RP-CG together (\ie, our proposed CG-Net), the model can increase mAP by at most $1.24\%$, in which some categories have large improvements, such as BD (Baseball diamond) $5.05\%$, SBF (Soccer-ball field) $3.14\%$, and RA (Roundabout) $2.77\%$. These results indicate that the feature presentation capabilities have been further improved by Base CG and RP-CG. As for the model efficiency, we can observe that Base CG and RP-CG respectively bring $0.59$ and $0.61$ M model $\#$Params with $51.53$ and $51.89$ GFLOPs. When these two models are deployed together, there is $1.79$ M model $\#$Params and $154.95$ GFLOPs increment. Our proposed CG is based on self-attention and calculates the similarity matrix between features so that GFlops increases from 289.26 to 444.21. In Table~\ref{t_multihead}, we compare the results of multi-head in our CG module and found that $\#$Params reduce $0.36$ M and mAP have $0.68\%$ increment when adding multi-head structure.


\myparagraph{(3) Different attention comparison}

In Table~\ref{t1}, we also show different attention mechanism comparison results, including Non-local~\cite{wang2018non} in spatial dimension and Squeeze-and-Excitation (SE) block from SENet~\cite{hu2018squeeze} in channel dimension. In more detail, We apply Non-local and SE blocks in different levels of the feature pyramid. We can observe that Non-local and SE block respectively bring $0.27\%$ and $0.13\%$ improvements for the bounding box mAP and improve $0.43\%$ mAP when combined together. When we apply the attention module in feature pyramid levels directly, improvements in mAP are less than our proposed CG module, and Non-local structure  also has higher computational complexity and model $\#$Params. From the table, we can observe that Non-local brings $2.18$ M model $\#$Params with $237.15$ GFLOPs. SE block has little change in $\#$Params and GFLOPs but improvements are very limited compared to the increase of mAP results from CG. When these two parts are deployed together, there is a $0.43\%$ mAP increment, less than Base CG $0.58\%$ and RP-CG $0.46\%$. 

\myparagraph{(4) Improving RPN input for aerial object detection} 

CG-Net shows significance when addressing complex background and worse imaging quality problems. Aerial images have complex geological structures, objects of different sizes, and object categories due to overhead shots from high altitudes, so they have a more complex background. In aerial object detection, worse imaging quality is detrimental to learning object features and directly affects model training. Therefore, we implement CG on pyramid features with Base CG and Rearrange Pyramid CG. In pyramid features, the size of proposals from the Region Proposal Network~(RPN)~\cite{ren2016faster} depends on the maximum response layer. Therefore, whether object proposals are selected accurately will affect the difficulty of the ROI module in training the detection box, which requires more accurate pyramid features. CG-Net can help the model learn richer communication information within and between each layer of pyramid features. To sum up, making Calibrated-Guidance operation for pyramid features is essential before input into region proposal network.
\begin{table*}[!htb]
\begin{center}
\scriptsize
\renewcommand\arraystretch{1.5} 
\setlength{\tabcolsep}{2pt}{ 
			\begin{tabular}{ccc|ccccccccccccccc|cc|c}
				\hline
				\hline
				Baseline  & Base CG & RP-CG & PL & BD & BR & GTF & SV & LV & SH & TC & BC & ST & SBF & RA & HA & SP & HC & GFLOPs / FPS & $\#$Params& mAP (\%)\\
				\hline
				\checkmark  & \xmark& \xmark& 88.53 & 77.70 & 51.59 & 68.80 & 74.02 & 76.85 & 86.98 & 90.24 & 84.89 & 77.68 & 53.91 & 63.56 & 75.88 & 69.48 & 55.50 &  289.26 / 20.9& 60.19 M & 73.06\\
				
				\checkmark & \checkmark & \xmark& 88.38	& 81.09 & 53.43 & 68.35 & 74.23 & 76.95 & 86.55 & 90.67 & 84.79 & \textbf{78.17} & 54.20 & 65.23 & 75.84 & 69.01 & 57.64 & 340.79 / 19.0 & 60.78 M& 73.64$_{\color{red}{ +0.58 }}$ \\
				
				\checkmark & \xmark& \checkmark & 88.49	& 77.18 & \textbf{54.02} & 68.85 & 74.49 & 76.67 & \textbf{87.09} & \textbf{90.79} & 86.17 & 77.58 & 54.54 & 65.03 & 75.87 & 69.11 & 56.94 & 341.15 / 17.2 &60.80 M & 73.52$_{\color{red}{ +0.46 }}$ \\
				
				\checkmark & \checkmark & \checkmark & \textbf{88.65}	& \textbf{82.75} & 53.02 & \textbf{69.65} & \textbf{74.77} & \textbf{77.48} & 86.99 & 90.32 & \textbf{86.38} & 77.23 & \textbf{57.05} & \textbf{66.33} & 75.46 & \textbf{70.22} & \textbf{58.23} & 444.21 / 13.4 & 61.98 M& \textbf{74.30}$_{\color{red}{ +1.24 }}$\\
				\hline
				Baseline  & Non-local & SE &  &  &  &  &  &  &  &  &  &  &  &  &  &  &  &  & & \\
				\hline
				\checkmark  & \checkmark& \xmark& 88.12 & 76.91 & 52.26 & 68.93 & 74.16 & 77.11 & 86.47 & 90.31 & 85.29 & 77.54 & 54.13 & 63.36 & 75.74 & 69.24 & 56.38 &  526.41 / 7.6& 62.37 M & 73.33$_{\color{red}{ +0.27 }}$\\
				
				\checkmark  & \xmark& \checkmark& 88.61 & 77.24 & 52.44 & 68.16 & 74.67 & 76.89 & 86.71 & 90.43 & 85.35 & 77.73 & 55.22 & 64.71 & 75.43 & 69.52 & 54.61 &  293.33 / 18.4& 60.25 M & 73.19$_{\color{red}{ +0.13 }}$\\
				
				\checkmark  & \checkmark& \checkmark& 88.58 & 79.42 & 53.47 & 68.84 & 74.32 & 77.13 & 87.05 & 90.61 & 84.89 & 77.87 & 56.26 & 64.96 & \textbf{75.91} & 69.48 & 57.37 &  533.94 / 7.2& 62.46 M & 73.49$_{\color{red}{ +0.43 }}$\\
				\hline
				\hline
			\end{tabular}
		}
	\end{center}
	\caption{Ablation study and different attention comparison on the test set of DOTA~\cite{xia2018dota} for oriented object detection. ResNet-101~\cite{he2016deep} is the backbone.}
	\label{t1}
\end{table*}

\begin{table*}[!htb]
\begin{center}
\scriptsize
\renewcommand\arraystretch{1.5} 
\setlength{\tabcolsep}{3pt}{ 
			\begin{tabular}{r|c|ccccccccccccccc|c}
				\hline
				\hline
				Methods & Backbone & PL & BD & BR & GTF & SV & LV & SH & TC & BC & ST & SBF & RA & HA & SP & HC & mAP (\%) \\
				\hline
				\multicolumn{18}{c}{{\textbf{Oriented object detection}}}\\
				\hline
				FR-O~\cite{xia2018dota} (CVPR 2018) & R-101 & 79.09 & 69.12 & 17.17 & 63.49 & 34.20 & 37.16 & 36.20 & 89.19 & 69.60 & 58.96 & 49.40 & 52.52 & 46.69 & 44.80 & 46.30 & 52.93 \\
				
				R-DFPN~\cite{yang2018automatic} (RS 2018)& R-101 & 80.92 & 65.82 & 33.77 & 58.94 & 55.77 & 50.94 & 54.78 & 90.33 & 66.34 & 68.66 & 48.73 & 51.76 & 55.10 & 51.32 & 35.88 & 57.94 \\
				
				R$^2$CNN~\cite{jiang2017r2cnn} (preprint 2017) & R-101 & 80.94 & 65.67 & 35.34 & 67.44 & 59.92 & 50.91 & 55.81 & 90.67 & 66.92 & 72.39 & 55.06 & 52.23 & 55.14 & 53.35 & 48.22 & 60.67 \\
				
				RRPN~\cite{ma2018arbitrary} (TMM 2018)& R-101 & 88.52 & 71.20 & 31.66 & 59.30 & 51.85 & 56.19 & 57.25 & 90.81 & 72.84 & 67.38 & 56.69 & 52.84 & 53.08 & 51.94 & 53.58 & 61.01 \\
				
				ICN~\cite{azimi2018towards} (ACCV 2018)& R-101 & 81.36 & 74.30 & 47.70 & 70.32 & 64.89 & 67.82 & 69.98 & 90.76 & 79.06 & 78.02 & 53.64 & 62.90 & 67.02 & 64.17 & 50.23 & 68.16 \\
				
    			RoI Trans~\cite{ding2019learning} (CVPR 2019)& R-101 & 88.64	& 78.52 & 43.44 & \textbf{75.92} & 68.81 & 73.68 & 83.59 & 90.74 & 77.27 & 81.46 & 58.39 & 53.54 & 62.83 & 58.93 & 47.67 & 69.56 \\
				
				CAD-Net~\cite{zhang2019cad} (TGRS 2019) & R-101 & 87.80 & 82.40 & 49.40 & 73.50 & 71.10 & 63.50 & 76.70 & \textbf{90.90} & 79.20 & 73.30 & 48.40 & 60.90 & 62.00 & 67.00 & 62.20 & 69.90 \\
				
				DRN~\cite{pan2020dynamic} (CVPR 2020)& H-104 & 88.91	& 80.22 & 43.52 & 63.35 & 73.48 & 70.69 & 84.94 & 90.14 & 83.85 & 84.11 & 50.12 & 58.41 & 67.62 & 68.60 & 52.50 & 70.70 \\
				
				O$^2$-DNet~\cite{wei2020oriented} (ISPRS 2020)& H-104 & 89.31	& 82.14 & 47.33 & 61.21 & 71.32 & 74.03 & 78.62 & 90.76 & 82.23 & 81.36 & 60.93 & 60.17 & 58.21 & 66.98 & 61.03 & 71.04 \\
				
				SCRDet~\cite{yang2019scrdet} (ICCV 2019) & R-101 & 89.98	& 80.65 & 52.09 & 68.36 & 68.36 & 60.32 & 72.41 & 90.85 & \textbf{87.94} & 86.86 & 65.02 & 66.68 & 66.25 & 68.24 & 65.21 & 72.61 \\
				
				CFC-Net~\cite{ming2021cfc} (TGRS 2021)& R-50 & 89.08	& 80.41 & 52.41 & 70.02 & 76.28 & 78.11 & 87.21 & 90.89 & 84.47 & 85.64 & 60.51 & 61.52 & 67.82 & 68.02 & 50.09 & 73.50 \\
				
				R$^3$Det~\cite{yang2019r3det} (AAAI 2021)& R-152 & 89.49	& 81.17 & 50.53 & 66.10 & 70.92 & 78.66 & 78.21 & 90.81 & 85.26 & 84.23 & 61.81 & 63.77 & 68.16 & 69.83 & 67.17 & 73.74 \\
				
				CSL~\cite{yang2020arbitrary} (ECCV 2020) & R-152 & \textbf{90.25}	& \textbf{85.53} & 54.64 & 75.31 & 70.44 & 73.51 & 77.62 & 90.84 & 86.15 & 86.69 & \textbf{69.60} & \textbf{68.04 }& 73.83 & 71.10 & 68.93 & 76.17 \\
				
				DAL~\cite{ming2020dynamic} (AAAI 2021)& R-50 & 89.69	& 83.11 & 55.03 & 71.00 & 78.30 & 81.90 & \textbf{88.46} & 90.89 & 84.97 & \textbf{87.46} & 64.41 & 65.65 & 76.86 & 72.09 & 64.35 & 76.95 \\
				
				R$^3$Det-DCL~\cite{yang2020dense} (CVPR 2021)& R-152 & 89.26 & 83.60 & 53.54 & 72.76 & \textbf{79.04} & 82.56 & 87.31 & 90.67 & 86.59 & 86.98 & 67.49 & 66.88 & 73.29 & 70.56 & 69.99 & 77.37\\
				\cdashline{1-18}[0.8pt/2pt]
				\textbf{Ours (CG-Net)} & R-101 & 88.75	& 85.18 & \textbf{57.41} & {71.88} & 73.23 & \textbf{82.68} & 88.14 & \textbf{90.90} & 86.00 & 85.37 & 62.99 & 66.74 & \textbf{77.98 }& \textbf{79.90} &\textbf{ 71.27} & \textbf{77.89}$_{\color{red}{ +0.52 }}$ \\
				\hline
				\multicolumn{18}{c}{\textbf{Horizontal object detection}}\\
				\hline
				SSD~\cite{liu2016ssd} (ECCV 2016)& R-101 & 44.74 & 11.21 & 6.22 & 6.91 & 2.00 & 10.24 & 11.34 & 15.59 & 12.56 & 17.94 & 14.73 & 4.55 & 4.55 & 0.53 & 1.01 & 10.94 \\

				YOLOv2~\cite{redmon2016you} (CVPR 2016) & R-101 & 76.90 & 33.87 & 22.73 & 34.88 & 38.73 & 32.02 & 52.37 & 61.65 & 48.54 & 33.91 & 29.27 & 36.83 & 36.44 & 38.26 & 11.61 & 39.20 \\
				
				R-FCN~\cite{dai2016r} (NIPS 2016)& R-101 & 79.33	& 44.26 & 36.58 & 53.53 & 39.38 & 34.15 & 47.29 & 45.66 & 47.74 & 65.84 & 37.92 & 44.23 & 47.23 & 50.64 & 34.90 & 47.24 \\
				
				FR-H~\cite{xia2018dota} (CVPR 2018)& R-101 & 80.32 & 77.55 & 32.86 & 68.13 & 53.66 & 52.49 & 50.04 & 90.41 & 75.05 & 59.59 & 57.00 & 49.81 & 61.69 & 56.46 & 41.85 & 60.46 \\
				
				FPN~\cite{lin2017feature} (CVPR 2017)& R-101 & 88.70	& 75.10 & 52.60 & 59.20 & 69.40 & 78.80 & 84.50 & 90.60 & 81.30 & 82.60 & 52.50 & 62.10 & 76.60 & 66.30 & 60.10 & 72.00 \\
				
				ICN~\cite{azimi2018towards} (ACCV 2018) & R-101 & 90.00	& 77.70 & 53.40 & \textbf{73.30 }& 73.50 & 65.00 & 78.20 & 90.80 & 79.10 & 84.80 & 57.20 & 62.10 & 73.50 & 70.20 & 58.10 & 72.50 \\
			    
			    YOLOv5~\cite{yolov5} (Github 2020) & D-53 & 81.13
                & 86.72 & 52.54 & 62.75 & 67.35 & 81.51 & 89.42 & 90.86 & 79.23 & 79.55 & 60.44 & 73.99 & 75.10 & 77.72 & 71.25 & 73.20 \\
			    
				SCRDet~\cite{yang2019scrdet} (ICCV 2019) & R-101 &\textbf{ 90.18} & 81.88 & 55.30 & 73.29 & 72.09 & 77.65 & 78.06 & \textbf{90.91} & 82.44 & \textbf{86.39} & \textbf{64.53} & 63.45 & 75.77 & 78.21 & 60.11 & 75.35 \\
				\cdashline{1-18}[0.8pt/2pt]
				\textbf{Ours (CG-Net)} & R-101 & 88.76 & \textbf{85.36} & \textbf{60.81} & 71.88 & \textbf{74.04 }& \textbf{83.43} & \textbf{88.29} & {90.89} & \textbf{86.00} & 85.55 & 63.51 & \textbf{67.02} &\textbf{ 78.78} & \textbf{80.86} & \textbf{68.70} & \textbf{78.26}$_{\color{red}{ +2.91 }}$\\
				\hline
				\hline
			\end{tabular}
		}
	\end{center}
	\vspace{-2mm}
	\caption{Result comparisons with state-of-the-art methods on the test set of DOTA~\cite{xia2018dota} for both oriented and horizontal object detection in aerial images. By ``Ours'' we mean that implementing Base CG and RP-CG on the baseline model at the same time. ``R-'' in the Backbone column denotes the ResNet~\cite{he2016deep}, ``D-'' in the Backbone column denotes the DarkNet~\cite{redmon2016you}, and ``H-'' denotes the Hourglass network~\cite{newell2016stacked}.}
	\vspace{-2mm}
	\label{t2}
\end{table*}

\begin{figure}[t]
\centering
\includegraphics[width=.48\textwidth,trim=270 100 400 170,clip]{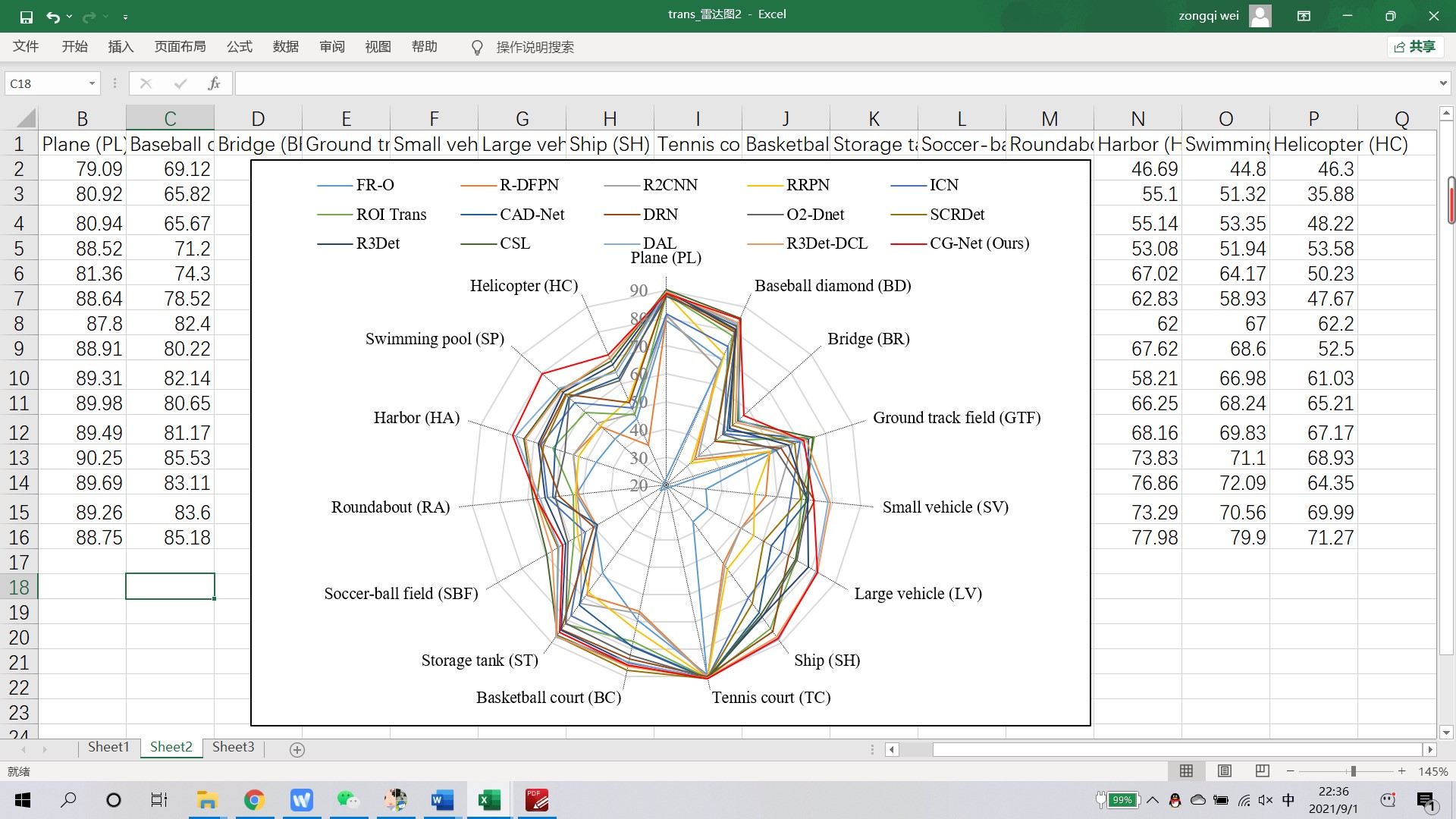}
\caption{Radar chart for the state-of-the-art methods on DOTA~\cite{xia2018dota} dataset for oriented object detection. Different colored lines represent different detectors. The larger areas enclosed by the outer line, the better recognition performance of the corresponding method. The value in this figure denotes the mAP.}
\label{Figure_radar2}
\end{figure}

\myparagraph{(5) Mismatching error rates on different scales} 

To reveal the effect of the proposed method on each level of the feature map, we define mismatching error rates on different scales in the feature pyramid, \ie, the selected level of each object is not consistent with the ground-truth level. It can be seen from Figure~\ref{Figure_histogram} that the mismatching error rate of each layer in the feature pyramid has been reduced after deploying our proposed method (\ie, the joint implementation of Base CG and RP-CG). Compared with the low-level feature in the feature pyramid that is more suitable for small objects, the reduction of error rates in high-level is obvious. For example, there are $0.1\%$, $0.2\%$, $0.1\%$, $0.7\%$, and $1.2\%$ error rate reduction from level $P2$ to $P6$. Therefore, the effectiveness of our method can be further confirmed. 

\myparagraph{\textbf{(6)} Visualized samples} 

From results of Ablation Experiment Table~\ref{t2} and Figure~\ref{Figure_view_all}, complex background and worse imaging quality, showing like Baseball diamond~(BD), Ground track field~(GTF), Plane~(PL) and Roundabout (RA), can be seen as obvious problems. Specifically, when detecting boxes are used to cover the whole objects, the boundary of boxes may show certain fuzziness, such as class Roundabout in Figure~\ref{Figure_view_new} left line 2, the problem of which is affected by complex background and labeling for completely covering object in aerial data. In left line 3, worse imaging quality leads to somewhat additional false detection boxes in local areas.

\begin{table}[t]
\begin{center}
\renewcommand\arraystretch{1.5} 
\setlength{\tabcolsep}{5pt}{ 
			\begin{tabular}{r|c|ccccccccccccccc|c}
				\hline
				\hline
				Methods & Backbone & mAP (\%) \\
				\hline
				R$^2$CNN~\cite{jiang2017r2cnn} (preprint 2017)& R-101 & 73.07\\
				RCI\&RC2~\cite{lb2017high} (ICPRAM 2017) & V-16 & 75.70\\
				RRPN~\cite{ma2018arbitrary} (TMM 2018)& R-101 & 79.08\\
				R$^2$PN~\cite{zhang2018toward} (GRSL 2018)& V-16 & 79.60\\
				RRD~\cite{liao2018rotation} (CVPR 2018)& V-16 & 84.30\\
				RoI Trans~\cite{ding2019learning} (CVPR 2019)& R-101 & 86.20\\
				Gliding Vertex~\cite{xu2020gliding} (TPAMI 2020) & R-101 & 88.20\\
				R-RetinaNet~\cite{lin2017focal} (ICCV 2017)& R-101 & 89.18\\
				R-RPN~\cite{li2020novel} (TGRS 2020)& V-16 & 89.20\\
				R$^3$Det~\cite{yang2019r3det} (AAAI 2021)& R-101 & 89.26\\
				RetinaNet-DAL~\cite{ming2020dynamic} (AAAI 2021)& R-101 & 89.77\\
				R$^3$Det-DCL~\cite{yang2020dense} (CVPR 2021)& R-101 & 89.46\\
				\cdashline{1-3}[0.8pt/2pt]
				\textbf{Ours (CG-Net)} & R-101 & \textbf{90.58}$_{\color{red}{ +1.12 }}$\\
				\hline
				\hline
			\end{tabular}
		}
	\end{center}
	\vspace{-2mm}
	\caption{Result comparisons with state-of-the-art methods on the test set of HRSC2016~\cite{lb2017high} for oriented object detection in aerial images. ``R-'' in the Backbone column denotes the ResNet~\cite{he2016deep}, and ``V-'' denotes the VGG network~\cite{simonyan2014very}. mAP is obtained on the VOC 2007 evaluation metric.}
	\label{t3}
\end{table}
\begin{figure}[t]
\centering
\includegraphics[width=.5\textwidth,trim=400 150 200 150,clip]{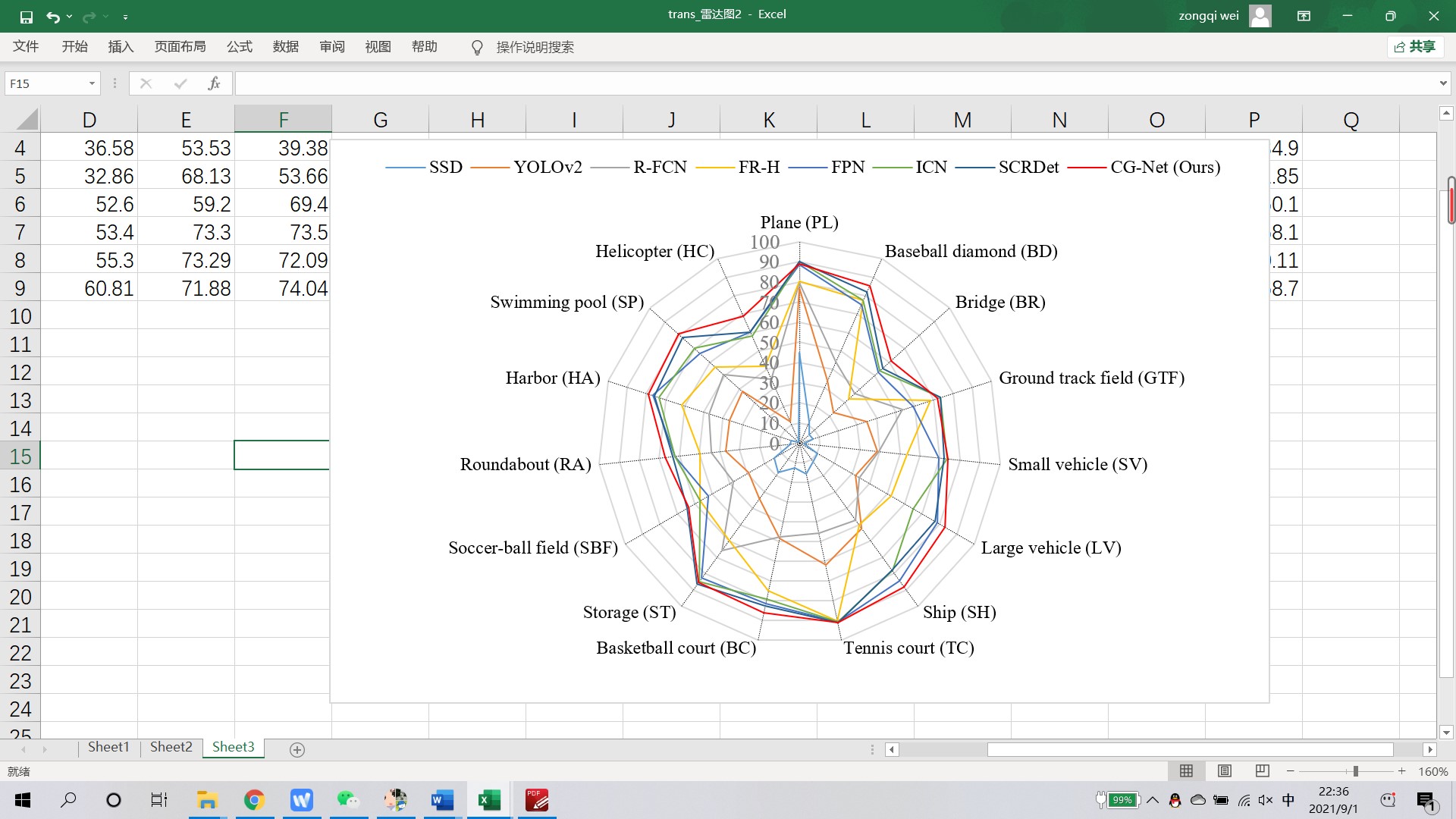}
\caption{Radar chart for the state-of-the-art methods on DOTA~\cite{xia2018dota} dataset for horizontal object detection. Different colored lines represent different detectors. The larger areas enclosed by the outer line, the better recognition performance of the corresponding method. The value in this figure denotes the mAP.}
\label{Figure_radar3}
\end{figure}
\subsection{Peer Comparisons}\label{sec4.3}
\textbf{On DOTA.} The experimental result on the test set of DOTA is shown in Table~\ref{t2}. The each-category mAP radar chart for oriented object detection is in Figure~\ref{Figure_radar2} and for horizontal object detection is in Figure~\ref{Figure_radar3}, to show the trend of the performance change. CG-Net achieves the best score among all compared methods, both on oriented object detection ($77.89\%$ mAP) and horizontal object detection ($78.26\%$ mAP). Inside $15$ categories, CG-Net achieves great results for oriented object detection ($6$ rank top) and horizontal object detection ($10$ rank top). It is worth noting that CG-Net utilizes a weaker backbone network to surpass the state-of-the-art by $0.52\%$ mAP on oriented object detection tasks (ResNet-152 \emph{vs} ResNet-101) and brings $2.91\%$ mAP increment for horizontal object detection with the same backbone. 
Compared to the approach (\ie, SCRDet~\cite{yang2019scrdet}) with the same backbone network (\ie, ResNet-101), our model has improved mAP by 5.82\%, which is quite remarkable in today's performance.
Rotating boxes avoid excessive background and clutter when calculating mAP compared with horizontal boxes so the improvements using our method for rotating boxes task are limited. While horizontal boxes contain more background, the features processed by our CG suppress background and highlight the object’s foreground features, so that mAP changes in horizontal boxes task are higher. Visualization results on the test set of DOTA are shown in Figure~\ref{Figure_view_all}. We can clearly observe that our model can achieve accurate recognition results.

\textbf{On HRSC2016.} From table~\ref{t3}, result comparisons with peer work on the test set of HRSC2016~\cite{lb2017high} show that the performance of our CG-Net surpasses the state-of-the-art methods by $90.58\%$ mAP, which increases $1.12\%$ mAP on the previous best model (R$^3$Det-DCL~\cite{yang2020dense}). Compared with the existing anchor strategy with large number and ratio, our CG-Net only combines original anchors setting with \{$1/2$, $1$, $2$\} ratio when training network, so it is worth noting how to utilize the presetting anchors to select or strengthen high-quality feature is reasonable and necessary considering efficiency and effectiveness. In addition, we also believe that our model can achieve further recognition performance with more complex aspect ratios.

\section{Conclusion and Future Work}
Complex background and worse imaging quality are obvious problems in aerial object detection. Most approaches tend to develop elaborate attention mechanisms for the space-time feature calibrations with arduous computational complexity. We have proposed a CG operation to enhance channel communications, which can determine the calibration weights for each channel. We implemented CG on the standard object detection backbone network with a feature pyramid network and we conducted extensive experiments on both oriented and horizontal object detection of aerial images. Experimental results on the challenging benchmarks indicated that the proposed CG-Net achieve state-of-the-art performance in accuracy with a fair computational overhead.
The each-category mAP radar chart for oriented object detection and horizontal object detection show the robust trend of its performance.  CG-Net surpasses the state-of-the-art for oriented object detection with a weaker backbone network (ResNet-101 \emph{vs} ResNet-152) and for horizontal object detection with the same backbone. We will explore to apply CG-Net to a broader range of natural scenes. Meanwhile, exploring how to use CG-Net in other visual tasks such as semantic segmentation and object re-identification is also an important direction.

\section*{Acknowledgement}
We thank the editor and reviewers for their careful and rigorous examination of our manuscript, which has benefited us a lot. This work was supported by AI+ Project of NUAA (XZA20003), Natural Science Foundation of China (62172212), Natural Science Foundation of Jiangsu Province (Grant BK20190065), Civil Aviation Administration of China (U2033202), the Free Exploration of Basic Research Project, the Local Science and Technology Development Fund Guided by the Central Government of China (No. 2021Szvup060).
\bibliographystyle{IEEEtran}
\bibliography{egbib}

\begin{IEEEbiography}[{\includegraphics[width=1in,height=1.25in,clip,keepaspectratio]{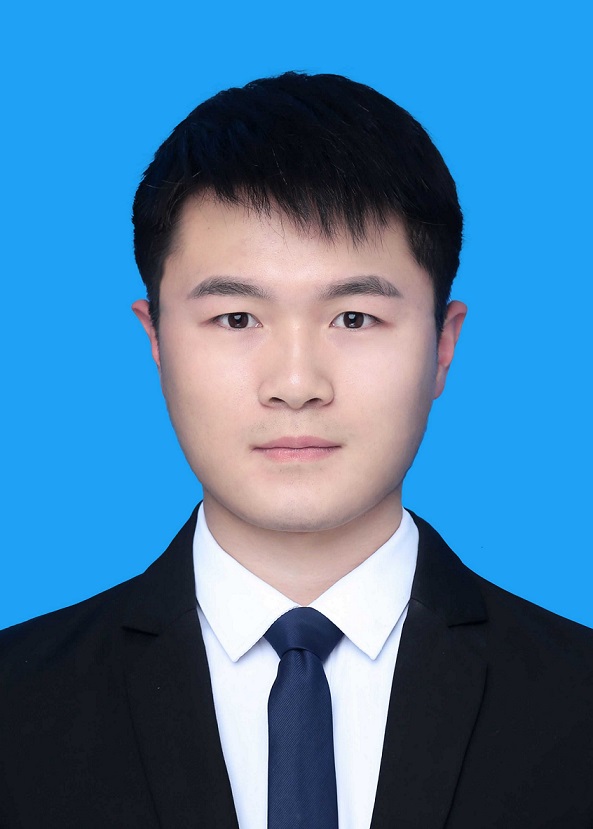}}]
{Zongqi Wei} received the BS degree in Computer Science and Technology at Nanjing Post and Telecommunication University, China, in 2019. He is now a master student with the College of Computer Science and Technology, Nanjing University of Aeronautics and Astronautics. His research interests include  computer vision, especially in object detection and video object segmentation. He is currently a research intern with TikTok, Beijing.
\end{IEEEbiography}
\begin{IEEEbiography}[{\includegraphics[width=1in,height=1.25in,clip,keepaspectratio]{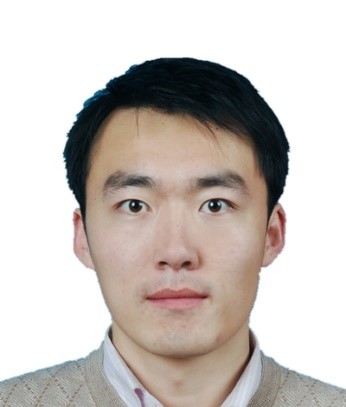}}]{Dong Liang}
received the B.S. degree in Telecommunication Engineering and the M.S. degree in Circuits and Systems from Lanzhou University, China, in 2008 and 2011, respectively. In 2015, he received Ph.D. at Graduate School of IST, Hokkaido University, Japan. He is currently an associate professor  with the College of Computer Science and Technology, Nanjing University of Aeronautics and Astronautics. His research interests include model communication in pattern recognition and image processing. He was awarded the Excellence Research Award from Hokkaido University in 2013. He has published several research papers including in Pattern Recognition, IEEE TIP/TCSVT/TGRS, and AAAI.
\vspace{-4mm}
\end{IEEEbiography}
\begin{IEEEbiography}[{\includegraphics[width=1in,height=1.25in,clip,keepaspectratio]{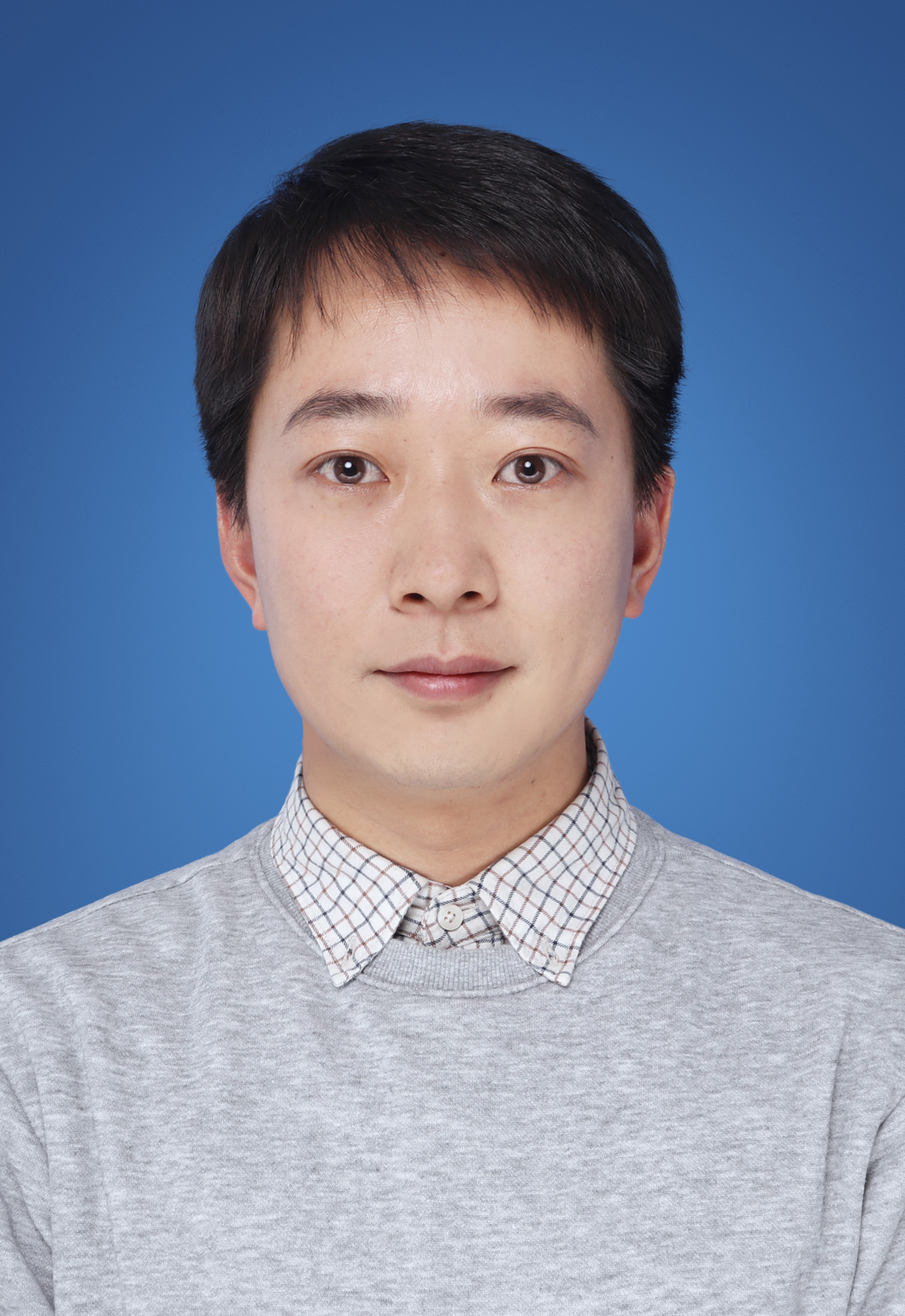}}]
{Dong Zhang} received his Ph.D. degree in Computer Science and Technology from Nanjing University of Science and Technology in 2022. He is now a postdoctoral research scientist at School of Engineering, The Hong Kong University of Science and Technology. His research interests include machine learning and computer vision, especially in object detection, semantic segmentation and video object segmentation. In these areas, he has published several journal and conference papers including IEEE Transactions on Cybernetics, Pattern Recognition, AAAI, IJCAI, ECCV, ICCV, and NeurIPS.
\end{IEEEbiography}
\begin{IEEEbiography}[{\includegraphics[width=1in,height=1.25in,clip,keepaspectratio]{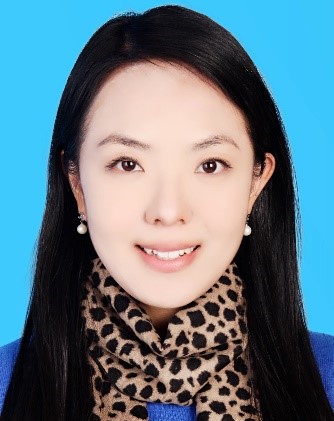}}]
{Liyan Zhang} received the Ph.D. degree in computer science from the University of California, Irvine, Irvine, CA, USA, in 2014. She is currently a Professor with the College of Computer Science and Technology, Nanjing University of Aeronautics and Astronautics, Nanjing, China. Her research interests include multimedia analysis, computer vision, and deep learning. Dr. Zhang received the Best Paper Award from the International Conference on Multimedia Retrieval (ICMR) 2013 and the Best Student Paper Award from the International Conference on Multimedia Modeling (MMM) 2016.
\end{IEEEbiography}

\begin{IEEEbiography}[{\includegraphics[width=1in,height=1.25in,clip,keepaspectratio]{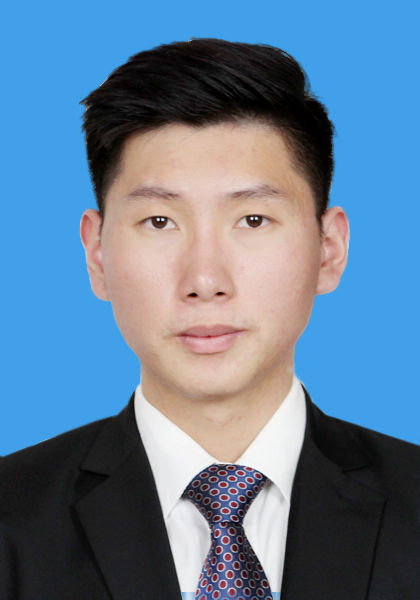}}]
{Qixiang Geng} received the BS degree in Computer Science and Technology at Nanjing Audit University in 2019. He is now a master student with the College of Computer Science and Technology, Nanjing University of Aeronautics and Astronautics. His research interests include computer vision, especially in object detection. He is currently a research intern with Senstime, Beijing.
\end{IEEEbiography}

\begin{IEEEbiography}[{\includegraphics[width=1in,height=1.25in,clip,keepaspectratio]{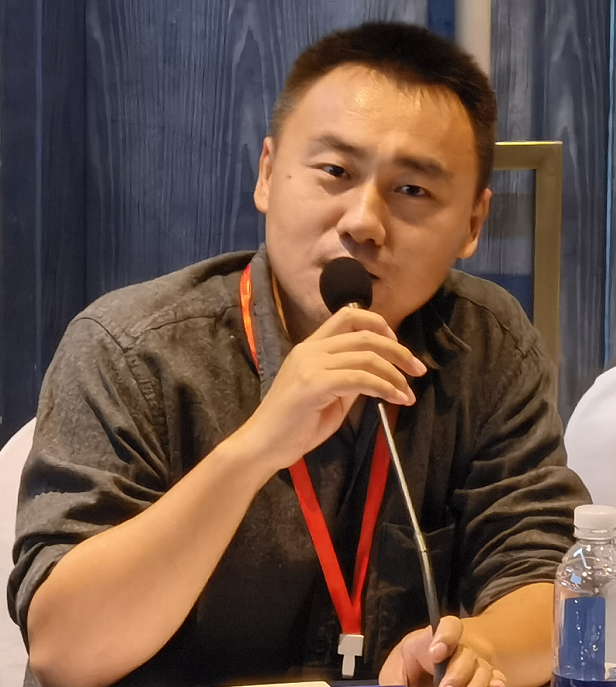}}]{Mingqiang Wei}
received his Ph.D degree (2014) in Computer Science and Engineering from the Chinese University of Hong Kong (CUHK). He is a professor at the School of Computer Science and Technology, Nanjing University of Aeronautics and Astronautics (NUAA). Before joining NUAA, he served as an assistant professor at Hefei University of Technology, and a postdoctoral fellow at CUHK. He was a recipient of the CUHK Young Scholar Thesis Awards in 2014. He is now an Associate Editor for the Visual Computer Journal, Journal of Electronic Imaging, Journal of Image and Graphics. His research interests focus on 3D vision, computer graphics, and deep learning.
\end{IEEEbiography}

\begin{IEEEbiography}[{\includegraphics[width=1in,height=1.25in,clip,keepaspectratio]{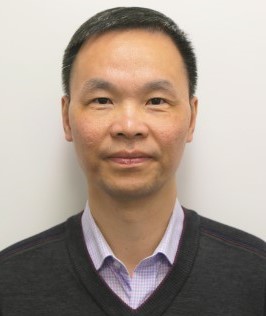}}]{Huiyu Zhou}
received a Bachelor of Engineering degree in Radio Technology from Huazhong University of Science and Technology of China (1990), and a Master of Science degree in Biomedical Engineering from University of Dundee of United Kingdom (2002), respectively. He was awarded a Doctor of Philosophy degree in Computer Vision from Heriot-Watt University, Edinburgh, United Kingdom (2006). Dr. Zhou currently is a full Professor at School of Informatics, University of Leicester, United Kingdom. He has published over 350 peer-reviewed papers in the field. 
\end{IEEEbiography}
\end{document}